\documentclass[journal,twoside]{IEEEtran}
\usepackage{amsmath,amssymb,amsfonts,bm}
\usepackage[colorlinks, linkcolor=magenta, anchorcolor=green, citecolor=blue, urlcolor=black]{hyperref}
\usepackage{hyperref}
\usepackage{algorithmic}
\usepackage{array}
\usepackage{textcomp}
\usepackage{stfloats}
\usepackage{url}
\usepackage{verbatim}
\usepackage{xcolor}
\usepackage{caption}
\usepackage{algorithmic}
\usepackage[linesnumbered,ruled]{algorithm2e}

\newcommand*{\circled}[1]{\lower.7ex\hbox{\tikz\draw (0pt, 0pt)%
    circle (.5em) node {\makebox[1em][c]{\small #1}};}}
\usepackage[square, numbers,sort&compress]{natbib}
\usepackage{bigstrut,multirow,rotating}
\usepackage{booktabs}
\usepackage{multirow}
\usepackage{array}
\usepackage{multirow}
\usepackage{rotating}
\usepackage{mathtools}
\usepackage{pifont}
\usepackage{mathrsfs}
\usepackage{colortbl}
\usepackage{graphicx,pifont}
\usepackage[normalem]{ulem}

\let\oldding\ding
\renewcommand{\ding}[2][1]{\scalebox{#1}{\oldding{#2}}}
\def\BibTeX{{\rm B\kern-.05em{\sc i\kern-.025em b}\kern-.08em
    T\kern-.1667em\lower.7ex\hbox{E}\kern-.125emX}}
 
\newcommand{\cmark}{\textcolor{black}{\ding{51}}}%
\newcommand{\xmark}{\textcolor{magenta}{\ding{55}}}%
\begin{document}

\title{GCoDE: Efficient Device-Edge Co-Inference for GNNs via Architecture-Mapping Co-Search}

\author{Ao~Zhou,~Jianlei~Yang,~\IEEEmembership{Senior~Member,~IEEE},~Tong~Qiao,~Yingjie~Qi,~Zhi~Yang,\\
~Weisheng~Zhao,~\IEEEmembership{Fellow,~IEEE}, Chunming Hu

\thanks{This work is supported in part by the National Natural Science Foundation of China (Grant No. 62072019), the Fundamental Research Funds for the Central Universities, the Beijing Natural Science Foundation (Grant No. L243031), and the National Key R\&D Program of China (Grant No. 2023YFB4503704 and 2024YFB4505601).
\textit{Corresponding authors is Jianlei Yang.}}
\thanks{A. Zhou and C. Hu are with School of Software, Beihang University, Beijing 100191, China.}
\thanks{J. Yang, T. Qiao, and Y. Qi are with School of Computer Science and Engineering, Beihang University, Beijing 100191, China, and Qingdao Research Institute, Beihang University, Qingdao 266104, China. Email: \url{jianlei@buaa.edu.cn}.}
\thanks{Z. Yang is with School of Computer Science and Engineering, Peking University, Beijing 100871, China.}
\thanks{W. Zhao is with School of Integrated Circuits and Engineering, Beihang University, Beijing 100191, China.}
\thanks{Manuscript received on September 2024, revised on August 2025, and accepted on October 2025.}
}

\maketitle
\begin{abstract}
Graph Neural Networks (GNNs) have emerged as the state-of-the-art graph learning method.
However, achieving efficient GNN inference on edge devices poses significant challenges, limiting their application in real-world edge scenarios.
This is due to the high computational cost of GNNs and limited hardware resources on edge devices, which prevent GNN inference from meeting real-time and energy requirements.
As an emerging paradigm, device-edge co-inference shows potential for improving inference efficiency and reducing energy consumption on edge devices.
Despite its potential, research on GNN device-edge co-inference remains scarce, and our findings show that traditional model partitioning methods are ineffective for GNNs.
To address this, we propose GCoDE, the first automatic framework for \underline{G}NN architecture-mapping \underline{Co}-design and deployment on \underline{D}evice-\underline{E}dge hierarchies.
By abstracting the device communication process into an explicit operation, GCoDE fuses the architecture and mapping scheme in a unified design space for joint optimization.
Additionally, GCoDE's system performance awareness enables effective evaluation of architecture efficiency across diverse heterogeneous systems.
By analyzing the energy consumption of various GNN operations, GCoDE introduces an energy prediction method that improves energy assessment accuracy and identifies energy-efficient solutions.
Using a constraint-based random search strategy, GCoDE identifies the optimal solution in $1.5$ hours, balancing accuracy and efficiency.
Moreover, the integrated co-inference engine in GCoDE enables efficient deployment and execution of GNN co-inference.
Experimental results show that GCoDE can achieve up to $44.9\times$ speedup and $98.2\%$ energy reduction compared to existing approaches across diverse applications and system configurations.
\end{abstract}

\begin{IEEEkeywords}
Graph Neural Networks, Device-Edge Co-Inference, Neural Architecture Search, Edge Devices, System Awareness
\end{IEEEkeywords}

\vspace{40mm}
\section{Introduction}\label{sec:introduction}

\IEEEPARstart{A}s edge devices become more intelligent and deep learning breakthroughs continue, the demand for deploying models to process various collected data in real-time on the device side is growing~\cite{sheng2022larger, xue2023sugar}.
However, limited hardware resources make it challenging to meet latency and energy requirements when deploying complex deep learning models~\cite{yan2020tinygnn}.
In particular, Graph Neural Networks (GNNs) have recently excelled in processing irregular data structures, making them a popular choice for graph-related applications in edge scenarios, such as point cloud processing~\cite{tailor2021towards} and natural language processing~\cite{chen2023multivariate}.
Additionally, the rising popularity of various sensors in mobile devices also encourages the deployment of GNNs to the wireless network edge for real-time sensing and interaction.
For instance, autonomous drones require immediate obstacle detection from point clouds~\cite{shi2020point}, where the latency of cloud communication is unacceptable. 
Likewise, executing speech-based interaction locally for smart assistants~\cite{dighe2020lattice} is crucial for safeguarding user privacy. 
However, the significant computational cost of GNNs and the limited hardware resources on edge devices pose major challenges to meeting these strict real-time and energy requirements, severely limiting their application in real-world edge scenarios.
This is demonstrated by deploying the popular point cloud processing model DGCNN~\cite{wang2019dynamic} on a Raspberry Pi 3B, which achieves less than $\textbf{0.3}$ fps, far below practical requirements (typically above $30$ fps~\cite{kalliomaki2019real}).

Research efforts have been made to address the inefficiency of GNNs on edge devices.
\cite{li2021towards, tailor2021towards} reduced GNN computation by manually simplifying the model structure.
Meanwhile, HGNAS~\cite{zhou2023hardware} and \cite{lu2022hardware} adopted a more efficient hardware-aware neural architecture search (NAS) approach to design hardware-friendly GNNs for edge devices.
Despite these performance improvements, GNN acceleration remains limited by constrained hardware resources.
For example, the hardware-efficient GNNs developed by HGNAS~\cite{zhou2023hardware} only increase point cloud processing speed to $2$ fps on the Raspberry Pi, as reported in their paper.
To address resource constraints, device-edge co-inference has emerged as a promising approach for deploying models at the network edge~\cite{ren2023survey, murshed2021machine}.
In this paradigm, choosing an appropriate split point to partition the model into device-side and edge-side execution parts can significantly reduce inference latency and on-device energy consumption.
However, most device-edge co-inference methods are designed for DNNs~\cite{li2023roulette, li2024optimal}, with few tailored to the unique computational patterns of GNNs.

To explore the device-edge co-inference paradigm for GNNs, Branchy-GNN~\cite{shao2021branchy} introduces a multi-branch design methodology that manually partitions the DGCNN model to minimize communication costs.
However, this approach neglects hardware characteristics and fails to explore novel architectures, resulting in limited performance gains.
In practice, simply partitioning existing models does not achieve the desired efficiency, and leveraging device-edge co-inference for GNNs remains challenging.
First, the intermediate data generated during GNN inference includes features and potentially graph-structured data, which must be carefully considered to balance communication and computation overheads.
Second, GNNs exhibit a hybrid computation pattern involving computation-intensive matrix operations and memory-intensive graph processing.
Each has distinct hardware sensitivities on heterogeneous devices, requiring an effective system-aware approach.
Additionally, separating architecture design from mapping often leads to sub-optimal performance, necessitating a co-optimized approach for both.
Furthermore, there is currently no effective way to perceive on-device energy consumption, which is critical for edge applications.
Therefore, an elegant co-inference methodology tailored for GNNs is needed.

\begin{figure}[t]
    \centering
    \includegraphics[width = 0.9\linewidth]{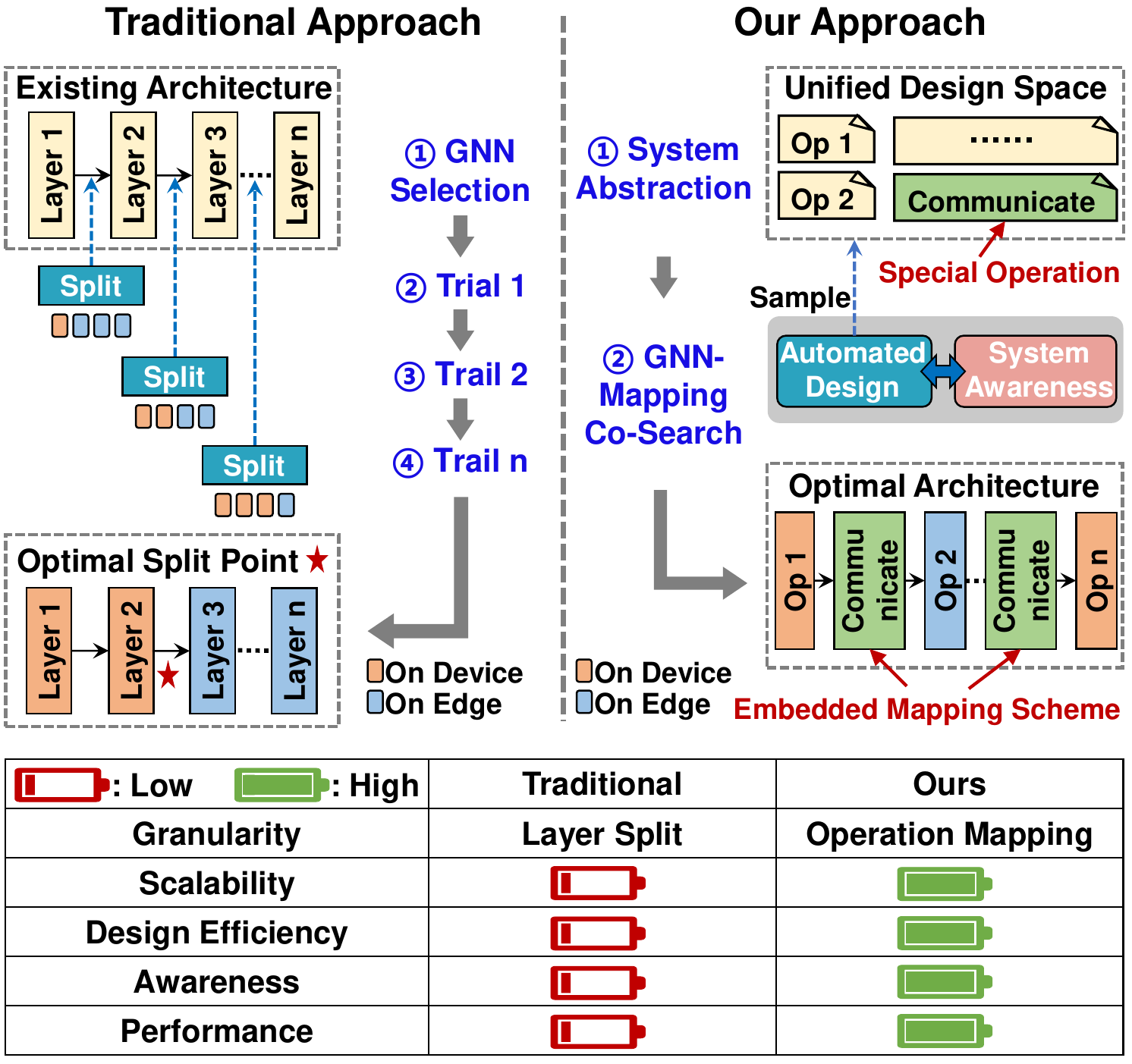}
    \caption{Device-Edge Co-Inference for GNNs.}
    \label{fig:approach_compare}
    \vspace{-12pt}
\end{figure}

To address the above-mentioned challenges, this paper proposes a novel NAS-based automated design and deployment framework for GNN device-edge co-inference, named GCoDE.
Given user requirements, GCoDE can efficiently search and deploy optimal GNN architectures with their concomitant mapping schemes for target systems, achieving both accuracy and efficiency under latency and energy constraints.
Specifically, GCoDE achieves joint optimization of GNN architectures and their operation mappings within the device-edge hierarchy in a single exploration.
This is based on an ingenious idea of \textbf{treating the inter-device communication process during co-inference as a special GNN operation} and incorporating it into the GNN design space.
Moreover, leveraging this system abstraction, an efficient GNN-based performance awareness approach is integrated to guide the constraint-based search process.
Furthermore, GCoDE develops a specialized device-edge co-inference engine for GNNs to support flexible operation mapping execution.
Fig.~\ref{fig:approach_compare} provides an intuitive comparison between traditional approaches and GCoDE. 
The \textit{scalability} refers to the ability of the framework to adapt and maintain performance across diverse system configurations, 
whereas \textit{design efficiency} denotes the reduction in manual effort and design cycle time. 
Unlike traditional methods that manually split existing models, GCoDE integrates automated design with system awareness to jointly optimize GNN architectures and operation mappings, 
thereby substantially enhancing scalability, design efficiency, and overall performance.

Our main contributions are summarized as follows:
\begin{itemize}
    \item We propose an architecture and operation mapping scheme co-search framework dubbed GCoDE which largely resolves the inefficiency of GNN device-edge co-inference. 
    To the best of our knowledge, GCoDE is the first automated design and deployment framework for GNNs in the device-edge co-inference paradigm, opening up an exciting perspective for exploring much more efficient GNN solutions.
    \item We propose a unified GNN design space for device-edge co-inference, enabling the joint optimization of architecture and operation mapping in a single constraint-based exploration, balancing accuracy, latency, and on-device energy consumption.
    \item To the best of our knowledge, GCoDE is also the first to achieve system performance awareness for GNNs in heterogeneous co-inference systems.
    The proposed latency and energy predictors achieve up to $85.3\%$ and $70.1\%$ accuracy within a $10\%$ error bound across different system configurations.
    \item We provide a comprehensive analysis of on-device energy consumption during GNN co-inference and extend the awareness dimensions to develop an energy prediction method that significantly improves accuracy over existing estimation methods.
    \item Extensive experiments on diverse applications and system configurations consistently validate the effectiveness of our GCoDE framework, e.g., GCoDE can achieve up to $44.9\times$, $21.6\times$ speedups and $98.2\%$, $96.2\%$ on-device energy savings over DGCNN and the existing SOTA GNN Device-Edge co-inference approach Branchy-GNN, respectively, while maintaining the same accuracy. 
\end{itemize}

The rest of the paper is organized as follows. 
Section~\ref{sec:works} discusses the related works. 
Section~\ref{sec:motivation} presents the motivations and provides an overview of the proposed framework. 
Section~\ref{sec:exploration} details the architecture–mapping co-exploration method, and Section~\ref{sec:awareness} describes the system performance awareness approach. 
Section~\ref{sec:deployment} introduces the device–edge deployment engine. 
Section~\ref{sec:experiment} reports the experimental results and analysis. 
Finally, Section~\ref{sec:conclusions} concludes the paper.

\section{Related Work}\label{sec:works}
In this section, we discuss the relevant prior work from three aspects.
The comparison of support features between GCoDE and other frameworks is illustrated in Tab.~\ref{tab:comparison}.

\begin{table}[t]
\centering
\caption{Comparison of support features}
\renewcommand\arraystretch{1.6}
\resizebox{1.0\linewidth}{!}{%
\begin{tabular}{|l|c|c|c|c|}
\hline
\textbf{Supported Features} & \textbf{GCoDE} & \textbf{HGNAS \cite{zhou2023hardware}} & \textbf{MaGNAS \cite{odema2023magnas}} & \textbf{Branchy \cite{shao2021branchy}} \\
\hline
Design Automation &   \cmark   & \cmark & \cmark & \xmark \\
\hline
Architecture Exploration & \cmark & \cmark & \cmark & \xmark \\
\hline
Performance Awareness & \cmark & \cmark & \cmark & \xmark \\
\hline
\quad $\rhd$ Single Device & \cmark & \cmark & \xmark & \xmark \\
\hline
\quad $\rhd$ Heterogeneous & \cmark & \xmark & \cmark & \xmark \\
\hline
\quad $\rhd$ Heterog. Wireless Edge & \cmark & \xmark & \xmark & \xmark \\
\hline
Multi-Objective Optimization & \cmark & \cmark & \cmark & \xmark \\
\hline
Device-Edge Deployment & \cmark & \xmark & \xmark & \cmark \\
\hline
Runtime Optimization & \cmark & \xmark & \xmark & \xmark \\
\hline
\end{tabular}
}
\label{tab:comparison}
 \vspace{-9pt}
\end{table}

\subsection{Graph Neural Networks}
GNN inference, based on the message-passing paradigm, typically involves two distinct operations: \textit{Aggregate} and \textit{Combine}~\cite{yan2020characterizing}.
The former aggregates features from source neighbors for each vertex, while the latter updates each vertex's feature using a weight matrix.
Additionally, for some edge applications, such as point cloud processing, the \textit{sample} operation is included to extract the graph structure from raw data for subsequent computation.
To accelerate GNN inference on edge devices, \cite{li2021towards} proposes reusing the \textit{sample} results across GNN layers, while \cite{tailor2021towards} introduces simplified message construction during neighbor aggregation.
These manual optimization efforts require extensive trial-and-error on the target device, resulting in weak scalability and lengthy design cycles.
Moreover, GraNNite~\cite{das2025grannite} focuses on hardware-aware optimization of GNN inference for resource-constrained NPUs in client devices through a multi-stage approach, achieving significant performance gains.
Unlike these works, which target single-device acceleration on specific hardware, GCoDE addresses the broader device–edge co-inference scenarios across heterogeneous environments by jointly optimizing GNN architectures and operation mappings with system performance awareness, enabling efficient and scalable GNN deployment at the wireless network edge.

\subsection{Device-Edge Co-Inference}

Device-edge co-inference is an emerging paradigm for edge AI applications that effectively addresses the constrained on-device resource and limited bandwidth present in device-only and edge-only inference~\cite{li2024optimal}.
The common approach to co-inference is to split a standard network into two parts and deploy them on a device and an edge server, respectively. 
Extensive studies have explored various splitting methods and collaborative mechanisms, greatly improving the inference efficiency of DNN models in edge applications like IoT~\cite{li2023roulette}. 
Nevertheless, the research for GNNs is still lacking.
Branchy-GNN~\cite{shao2021branchy} (denoted as Branchy) was the first co-inference method for GNNs, improving inference efficiency by manually splitting the existing model. 
Unfortunately, the lack of hardware awareness and separate design manner prevented it from realizing the full potential of this paradigm.
In contrast, GCoDE integrates system performance awareness with architecture-mapping co-search, ensuring adaptation to various system configurations.

\subsection{Graph Neural Architecture Search}
GNN NAS has been proposed to design application-customized GNNs, addressing the inefficiencies of manual design~\cite{gao2021graph}. 
Most early GNN NAS research focused on enhancing model expressiveness~\cite{cai2021rethinking}, neglecting inference efficiency.
To meet real-time edge application needs, hardware-aware GNN NAS has emerged to optimize both accuracy and efficiency.
MaGNAS~\cite{odema2023magnas} presents a hardware-aware GNN NAS framework that uses a lookup table (LUT) to guide GNN search on the MPSoC platform.
HGNAS~\cite{zhou2023hardware} integrates a GNN-based hardware performance predictor into its NAS framework, achieving SOTA performance on various edge devices. 
However, these existing GNN NAS frameworks focus solely on single-device inference and do not effectively handle heterogeneous device-edge hierarchies in wireless network environments.
GCoDE's system performance awareness is explored from a new perspective by abstracting the co-inference process into a unified GNN design space that naturally covers heterogeneous hardware sensitivities and network conditions.

\section{GCoDE: Motivation \& Overview}\label{sec:motivation}

\subsection{Why GNN Co-Inference is Inefficient}\label{sec:why_inefficient}

Below, We analyze the inefficiency bottlenecks in GNN co-inference through three observations using DGCNN~\cite{wang2019dynamic} as the examined GNN model.

\begin{figure}[t]
    \centering
    \includegraphics[width = 0.8\linewidth]{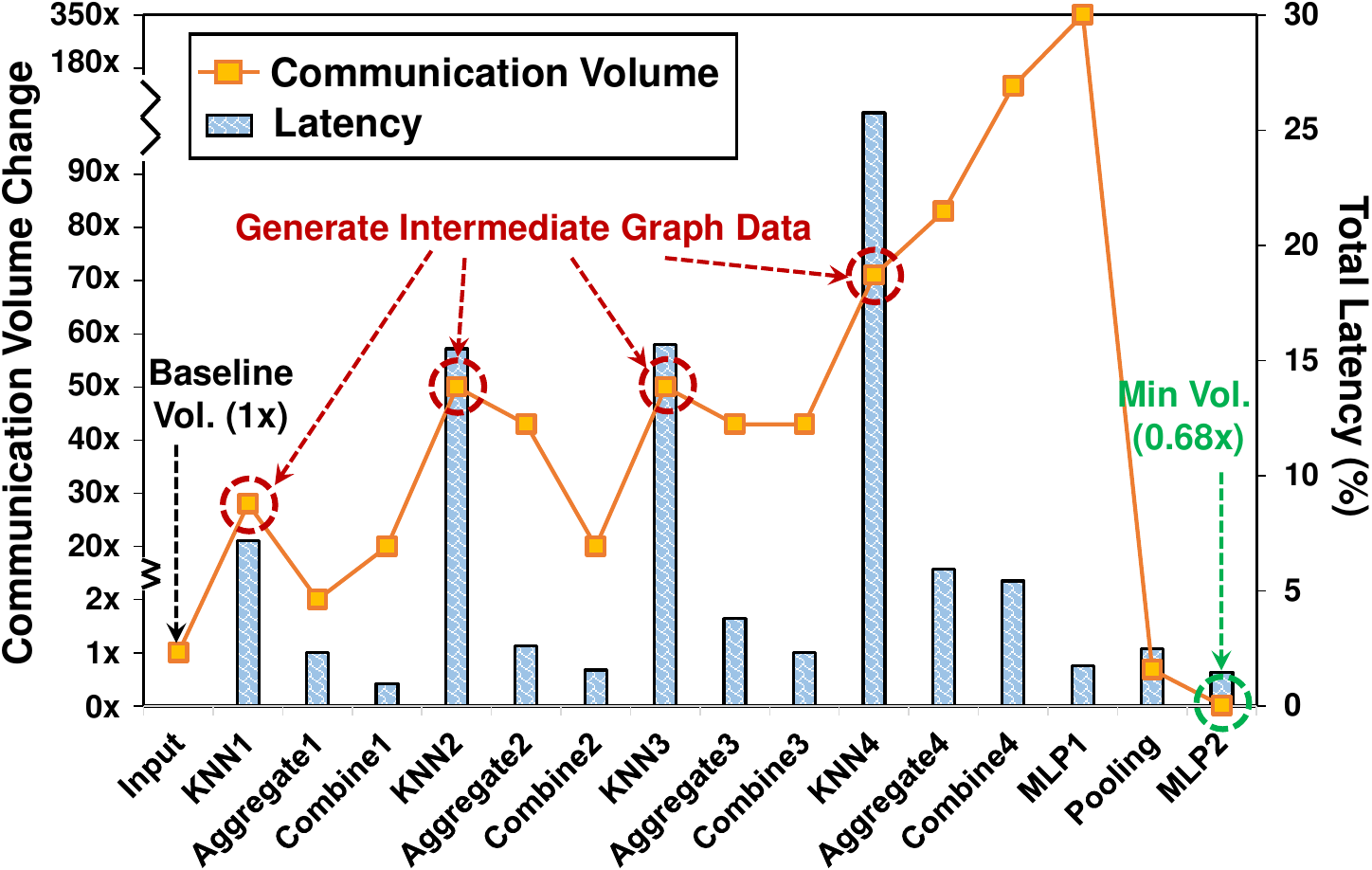}
    \caption{Changes in the communication volume, transferred between operations, and the percentage of total latency for each operation in DGCNN on Jetson TX2.}
    \label{fig:ob1}
    \vspace{-6pt}
\end{figure}

\textbf{Observation \raisebox{-0.1em}{\ding[1.3]{182}}: The trade-off between computation and communication is difficult to manage.}
As shown in Fig.~\ref{fig:ob1}, the communication volume refers to the size of intermediate data—such as node features and, in some cases, graph structures—transferred between operations. 
For instance, the \textit{KNN} operation can produce graph-structured intermediate data, significantly increasing the communication overhead when split after it.
Additionally, it can be observed that as the feature dimension expands with more GNN layers, the \textit{KNN} operation becomes increasingly time-consuming.
Moreover, the split point that requires minimal intermediate data transmission is at the last \textit{MLP}, which assigns most of the computation to the edge device.
Since edge devices generally have much lower computational capability and fewer resources than edge servers, excessive computation on the device, especially for weaker platforms such as the Raspberry Pi, can cause severe on-device workload. 
This situation leads to underutilization of edge resources and ultimately degrades the overall inference performance.
Thus, designing an effective device-edge co-inference solution for GNNs suffers from a difficult trade-off between communication and computation.

\begin{figure}[t]
    \centering
    \includegraphics[width = 0.8\linewidth]{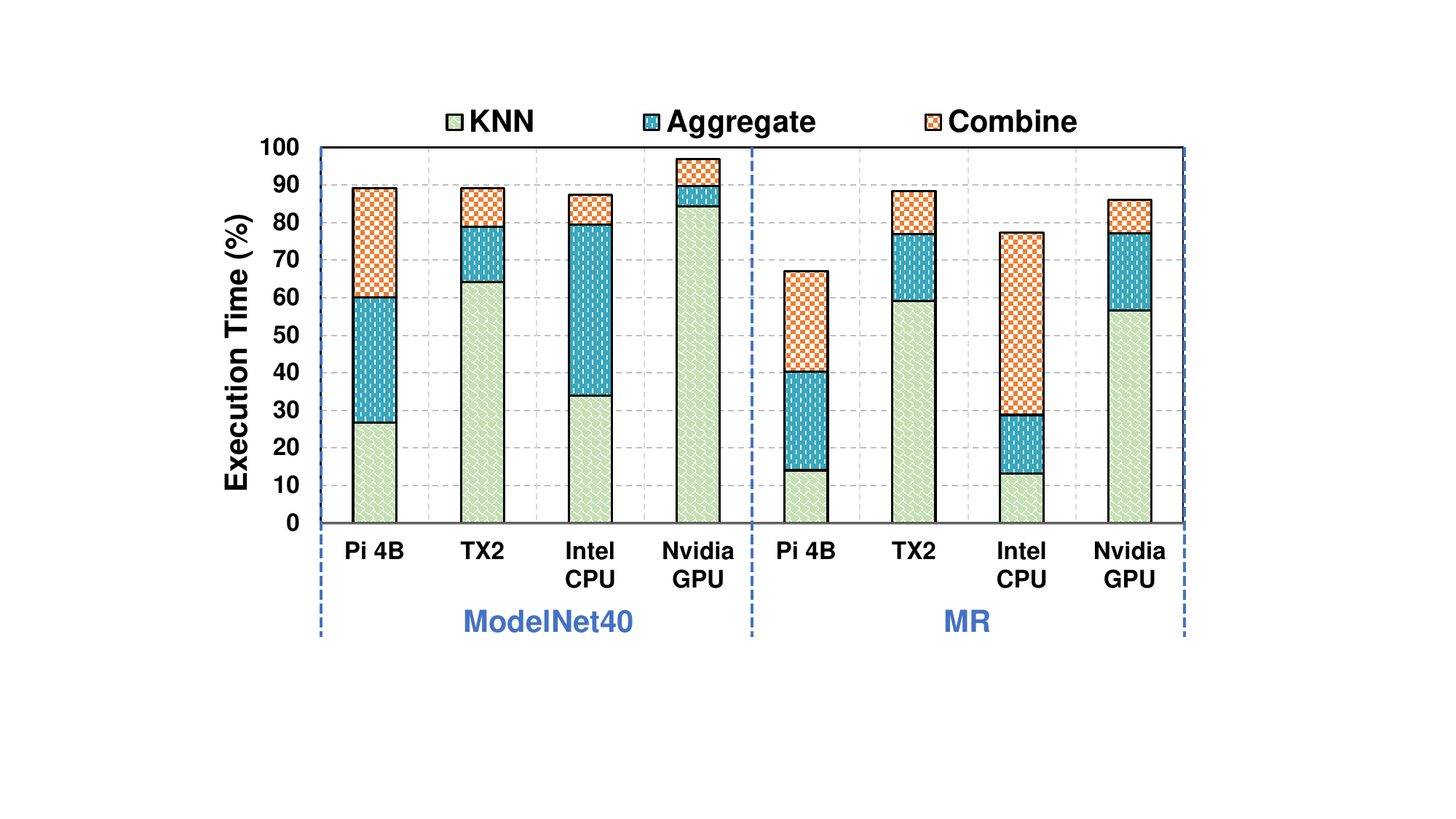}
    \caption{Execution time breakdown of DGCNN across various devices and datasets.}
    \label{fig:ob2}
    \vspace{-9pt}
\end{figure}

\textbf{Observation \raisebox{-0.1em}{\ding[1.3]{183}}: The hardware sensitivity of GNNs varies across heterogeneous hierarchies.}
Fig.~\ref{fig:ob2} shows the varying hardware sensitivities of GNN operations, highlighting the need for heterogeneity awareness in design.
For the point cloud dataset ModelNet40~\cite{wu20153d}, \textit{KNN} operation is the main execution bottleneck on the Jetson TX2 and Nvidia GPU.
This is because the parallel processing capability of these devices is hindered by the intensive and irregular memory accesses of the \textit{KNN} operation.
Additionally, the \textit{Aggregate} operation becomes the main bottleneck on the Intel CPU, while on the Raspberry Pi, all operations are time-consuming due to resource constraints.
Unlike point cloud data, the text dataset MR~\cite{ZhangYCWWW20} features fewer nodes ($1024$ vs. $17$) and larger feature dimensions ($3$ vs. $300$), resulting in different execution characteristics.
For example, the \textit{Combine} operation became a computational bottleneck on the Intel CPU.
This analysis shows that each GNN operation suits different platforms, explaining why traditional layer-level partitioning methods are inefficient.

\begin{figure}[t]
    \centering
    \includegraphics[width = 0.8\linewidth]{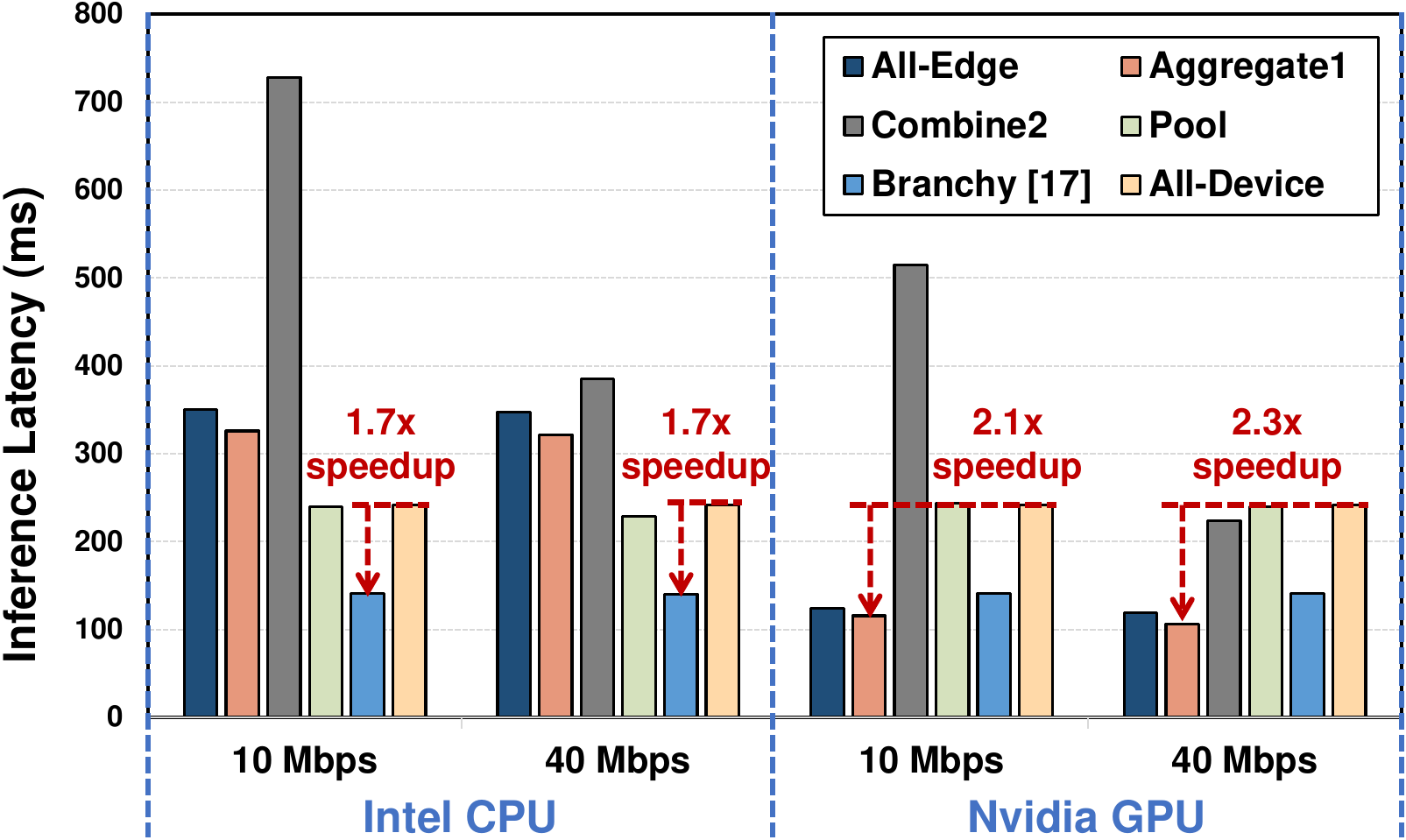}
    \caption{Performance of various potential splitting schemes on DGCNN, with Jetson TX2 serving as the device.}
    \label{fig:ob3}
    \vspace{-9pt}
\end{figure}

\textbf{Observation \raisebox{-0.1em}{\ding[1.3]{184}}: Detachment between architecture design and mapping design hinders system performance.}
Most traditional methods manually select the optimal split point within existing architectures to deploy GNNs to device-edge co-inference systems, such as Branchy-GNN~\cite{shao2021branchy}.
Fig.~\ref{fig:ob3} shows the performance of deploying DGCNN on device-edge systems based on various potential collaborative schemes.
Considering that the NVIDIA 1060 GPU offers roughly $3\times$ the computational capability of the Jetson TX2, their collaborative execution could intuitively achieve around a $4\times$ speedup even without further optimizations. 
This indicates that the $2.3\times$ improvement achieved by the prior method does not fully exploit the potential of collaboration and still leaves room for further optimization.
The key issue of these inefficient results lies in the lack of co-optimization between the GNN architecture and the operation mapping scheme.

\begin{figure*}[t]
    \centering
    \includegraphics[width = 1\linewidth]{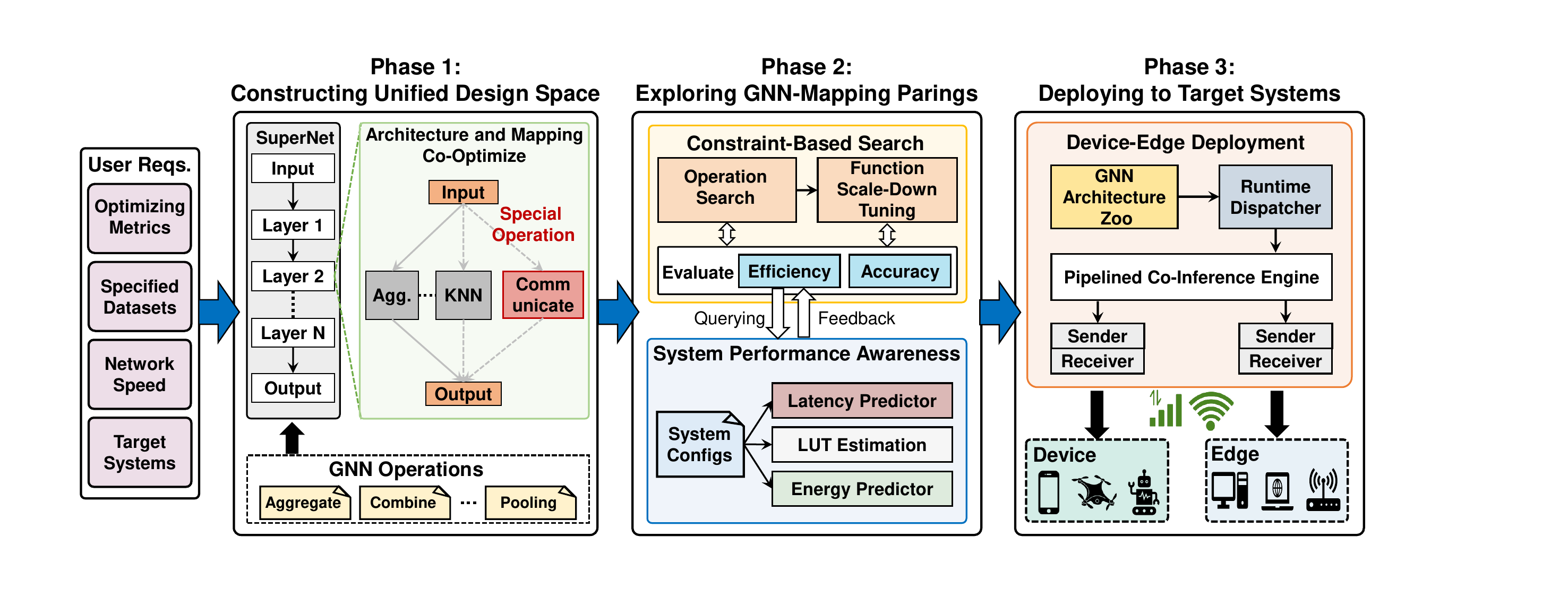}
    \caption{Overview of GCoDE framework.}
    \label{fig:framework}
    \vspace{-6pt}
\end{figure*}

\subsection{Why GCoDE Boosts System Efficiency}

Fig.~\ref{fig:framework} shows an overview of the proposed GCoDE framework, which cleverly resolves the above dilemma.
We first leverage a unified GNN design space construction method tailored for the co-inference paradigm to perfectly integrate the mapping scheme and GNN architecture into the same design space.
Then, an efficient NAS approach is utilized to co-optimize the GNN architecture along with its embedded mapping scheme within this specialized design space for multiple objectives (e.g., accuracy, latency, and energy).
The performance of candidate architectures during the exploration process is evaluated using system performance awareness methods, guiding the search toward more efficient solutions.
Finally, leveraging the integrated GNN-specific deployment approach, efficient GNN co-inference is achieved.
The motivating intuition is very simple: The unique aspect of device-edge co-inference is the communication process, and if we abstract this communication process as a special GNN operation, the sampled architecture from such a unified design space will inherently embed the mapping scheme.
Consequently, this system-aware exploration of the unified design space can largely resolve the aforementioned dilemma.
In this way, the designed collaborative solution evolves from a simple model split point to a novel GNN architecture with fine-grained operation mapping.
Ultimately, GCoDE achieves optimal alignment among the system, GNN architecture, and mapping, significantly enhancing the efficiency.

\section{Architecture-Mapping Co-Exploration}\label{sec:exploration}

\subsection{Problem Formulation}
Leveraging our elegant abstraction of the co-inference process, collaborative schemes are inherently present in GNN architectures.
Thus, co-exploring architectures and mappings reduces to exploring specialized GNN architectures with embedded mappings.
Specifically, this work aims to co-optimize the accuracy and efficiency of GNNs deployed on device-edge co-inference systems.
Given user requirements: edge device $\mathcal D$, edge server $\mathcal E$, network speed $\mathcal S$, latency constraint $\mathcal{C}_{lat}$ and on-device energy constraint $\mathcal{C}_{e}$, the optimization process can be formulated as:
\begin{equation*}
 \arg \mathop {\max }\limits_{{\alpha \in \mathbb{A}} }  \left( {acc_{val}}\left({{\mathcal W}^ * }, \alpha \right) - \lambda {\mathcal{P}_{sys}({\alpha}, \mathcal{D}, \mathcal{E}, \mathcal{S})} \right),
\end{equation*}
\begin{equation*}
    \begin{split}
            s.t. \quad & {{\mathcal W} ^ * } = \arg \mathop {\max }\limits_{\mathcal W}\left\{{acc_{train}} \left( {\mathcal W}, {\alpha} \right)\right\} \\
            & {{\mathcal{L}_{sys}} < \mathcal{C}_{lat}} \quad and \quad E_{dev} < \mathcal{C}_{e}
    \end{split} \quad,
\end{equation*}
where $\mathcal{W}$ denotes the model weights, $acc_{train}$ represents the training accuracy, $acc_{val}$ is the validation accuracy, $\alpha$ refers to the architecture selected for optimization from the unified GNN design space $\mathbb{A}$, $\mathcal{P}_{sys}$ indicates system performance, including inference latency $\mathcal{L}_{sys}$ and on-device energy consumption $E_{dev}$, and $\lambda$ is a scaling factor that balances the trade-off between accuracy and efficiency.
Note that system performance is jointly determined by the architecture, device-edge configurations, and network conditions.

\subsection{Unified GNN Design Space for Co-Inference}\label{sec:space}

\begin{figure}[t]
    \centering
    \includegraphics[width = 1.0\linewidth]{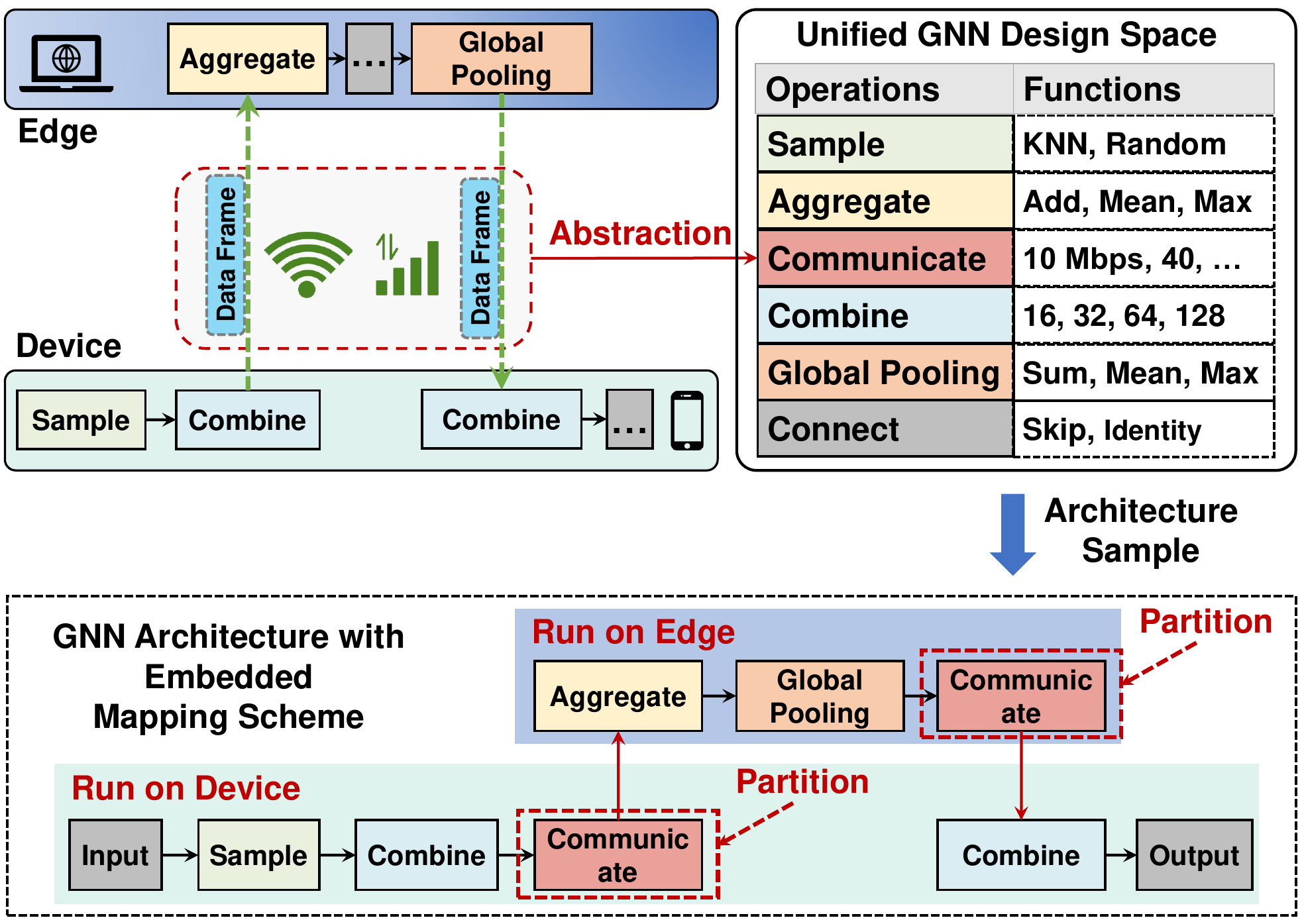}
    \caption{Unified GNN design space for architecture design and operation mapping in device-edge co-inference.}
    \label{fig:mapping}
    \vspace{-6pt}
\end{figure}

Fig.~\ref{fig:mapping} illustrates the unified design space for GNN device-edge co-inference, which supports the joint optimization of architectures and operation mappings.
This is based on a novel concept: communication between the device and the edge can be treated as a specialized operation within the GNN architecture.
As observed, the main difference between co-inference and on-device inference is that GNN operations are executed on different devices, using network communication to transfer intermediate data.
Thus, we abstract \textit{communicate} as a specialized GNN operation, constructing a unified design space for co-inference.
In this way, GNN architectures sampled from this design space inherently incorporate operation mapping schemes across device-edge hierarchies.
Consequently, joint optimization of architectures and collaborative schemes reduces to optimizing GNN architectures within the unified design space.
Each \textit{communicate} operation in the sampled GNN architecture represents a partition point, where the next operation is executed on the opposite side of the device-edge hierarchy.
By integrating architecture and mapping design, GCoDE can explore more flexible collaboration patterns instead of merely seeking a split point.
Furthermore, this fine-grained operation mapping enables the system to leverage device heterogeneity, maximizing each device's capabilities.

To utilize the one-shot NAS approach and avoid the overhead of retraining sub-architectures, GCoDE structures the unified design space $\mathbb{A}$ as a supernet, comprising six operations per layer: \textit{Sample}, \textit{Aggregate}, \textit{Communicate}, \textit{Combine}, \textit{Global Pooling}, and \textit{Connect}, each having distinct functional properties, as shown in Fig.~\ref{fig:mapping}.
During supernet training, linear layers are employed to align the dimensions of all operations within each layer, and these layers are removed before the search to maintain efficiency.

\begin{algorithm}[t]
\small
\KwIn{Target device-edge co-inference system ${Sys}$, system constraints: $\{\mathcal{C}_{lat}, \mathcal{C}_{e}\}$, predefined function setting $f$, max operation search and function tuning iteration:{ $T$, $T_{f}$}.}
\KwOut{The best found GNN architectures $\alpha^*$.}
Initialize $\alpha^* \gets \emptyset$, $Ops \gets \emptyset$, $func \gets \emptyset$.\\
Pre-train GNN supernet $\mathcal{N}_{super}$ with $f$.\\   
{/* Stage 1: operation search */} \\
\For{$1 \leq t \leq T$}{
    \While{\textnormal{Check($Ops$)}}{
        $Ops \gets \textnormal{Random}\left(\mathbb{A}\right)$ \hfill{// Sample valid operations}\\
    }
    $\alpha = \mathcal{N}_{super}(Ops,f)$  \hfill{// Construct architecture}\\
    $\mathcal{P}_{sys} \gets \textnormal{Evaluate}\left(Sys, \alpha \right)$ \hfill{// Evaluate performance}\\
    \eIf{$\mathcal{L}_{sys} < \mathcal{C}_{lat}$ \textnormal{and} $E_{dev} < \mathcal{C}_{e}$}{
        $score \gets \left( acc_{val} - \lambda \mathcal{P}_{sys}\right)$ \hfill{// Full evaluation}
    }{$score \gets (-1)$ \hfill{// Discard failed architectures}}
    $\alpha^*$ $\gets$ \textnormal{Update($Ops$, $f$, $score$)} \hfill{// Update operations}\\
}
{/* Stage 2: function scale-down tuning */} \\
\For{$1 \leq t \leq T_{f}$}{
$f' \gets \textnormal{Random}(\mathbb{A}, f)$ \hfill{// Scale-down functions}\\
$\alpha^*$ $\gets$ \textnormal{Update($Ops$, $f'$, $acc_{val}$)} \hfill{// Update functions}\\
}
\textbf{return} $\alpha^*$ \hfill{// Top-performing designs}
\caption{\small Constraint-based search strategy.}
\label{gnnalgo}
\end{algorithm}

\subsection{Constraint-Based Search Strategy}\label{sec:search}
While the unified design space offers benefits, it also introduces complexity to the exploration process.
Each architecture sampled from this design space may include invalid configurations, such as consecutive \textit{communicate} operations that are redundant and introduce unnecessary communication overhead.
In such situations, intelligent algorithms, such as evolutionary algorithms (EA), are likely to struggle with identifying valid architectures instead of pursuing more efficient ones (see Sec.~\ref{sec:ablation}).
When dealing with such complex design space, simpler random search strategies may lead to surprising outcomes and offer the potential for better optimization in a shorter duration~\cite{yu2019evaluating}.
Additionally, random search offers greater customizability and flexibility, enabling the discovery of multiple optimal solutions for different objectives at minimal cost.
Thus, GCoDE integrates a constraint-based random search strategy to boost exploration efficiency, as depicted in Alg.~\ref{gnnalgo}.

Specifically, the search process consists of two stages: operation search and function scale-down tuning.
Given the system configuration ${Sys}$ and application requirements $\{\mathcal{C}_{lat}, \mathcal{C}_{e}\}$, GCoDE initially establishes an appropriate function setting $f$ for the GNN supernet $\mathcal{N}_{super}$, referring to the target GNN model.
Next, GCoDE trains the supernet, focusing on accuracy, to produce shared weights for further operation search.
During the operation search, architecture validity is checked at each sampling step to avoid the overhead of evaluating invalid architectures.
Subsequently, multi-objective optimization is conducted with a focus on achieving optimal system performance $\mathcal{P}_{sys}$ and validation accuracy $acc_{val}$.
Finally, the function settings of the candidate architectures are further optimized during the function scale-down tuning stage to obtain the optimal GNN architecture.
The details of the search process are outlined below.

\textbf{Stage 1: Operation search.}
This stage aims to identify optimal operation settings $Ops$ that allow candidate architecture $\alpha$ to achieve higher scores while meeting validity and performance constraints.
Each candidate architecture is generated through random sampling of operations at each layer of the supernet.
The validity of the candidate operation set is first evaluated using a checking function that identifies common invalid architecture configurations.
System performance metrics, including latency $\mathcal{L}_{sys}$ and on-device energy consumption $E_{dev}$, are assessed only on valid architectures to ensure compliance with user requirements.
In this manner, substandard candidates are quickly pruned, significantly improving exploration efficiency.
The system performance of candidate architectures is evaluated using our unique performance-awareness method, reducing the overhead of real-time measurements.
Subsequently, architectures meeting the criteria are evaluated for task accuracy using the shared weights from the pre-trained supernet, without any retraining. This follows the one-shot NAS paradigm, which avoids retraining for each candidate and significantly improves the search efficiency.
Finally, the optimal architectures $\alpha^*$ are updated by scoring the candidate operation set on accuracy and system performance metrics.
Note that by leveraging the unified design space, the device–edge mapping is inherently embedded in each sampled architecture.
This eliminates the need for a separate partitioning step and enables joint optimization of architecture and mapping in a single search process.

\begin{figure}[t]
    \centering
    \includegraphics[width = 1.0\linewidth]{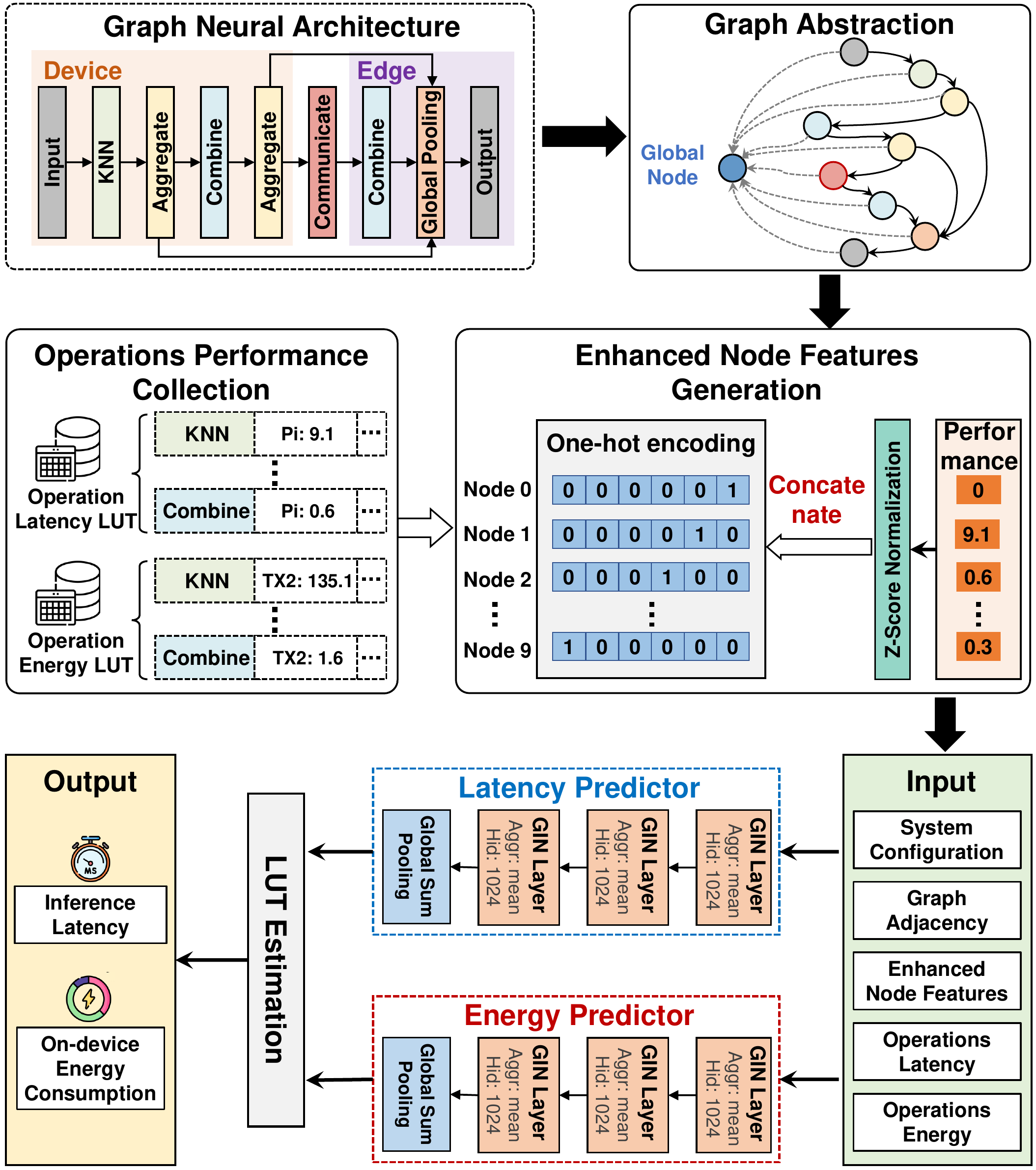}
    \caption{Latency and on-device energy consumption prediction for GNNs on device-edge hierarchies.}
    \label{fig:predictor}
    \vspace{-12pt}
\end{figure}

\textbf{Stage 2: Function scale-down tuning.}
In this stage, GCoDE aims to identify function settings that maximize efficiency while maintaining acceptable accuracy loss.
For example, by reducing feature dimensions in the \textit{Combine} operations of candidate architectures to improve computational efficiency.
As scaling down the function inevitably reduces computation, this phase focuses solely on evaluating the accuracy of the candidate architecture.
Additionally, changes to function settings require retraining the sub-architecture, as supernet weights are not reusable, necessitating a balance between design duration and performance.
Given that the candidate architecture from the operation search stage already satisfies requirements for accuracy, latency, and energy consumption, this stage is optional.

\section{System Performance Awareness}\label{sec:awareness}

To mitigate the considerable overhead from real-time measurements, GCoDE utilizes an efficient performance awareness method, as demonstrated in Fig.~\ref{fig:predictor}, to evaluate the performance of candidate GNN architectures deployed on the target device-edge co-inference system.
Specifically, we focus on two crucial performance metrics in edge applications: inference latency and on-device energy consumption.
Additionally, the implementation of the proposed system performance awareness method, including training scripts, model architecture, and dataset preparation guidelines, is publicly available at \url{https://github.com/BUAA-CI-LAB/GCoDE-Predictor}.

\subsection{Problem Simplification}
The system performance awareness problem is greatly simplified by the unified GNN design space approach.
Specifically, GNN architectures sampled from this space inherently contain implicit mapping schemes, heterogeneous device information, and network information.
Thus, the challenge of assessing the performance of GNNs deployed on a heterogeneous device-edge co-inference system can be reframed as the task of extracting information from the GNN architecture.
By abstracting the GNN architecture into an architecture graph, the problem is ultimately simplified to learning this architecture graph.
Since GNNs excel at handling such graph-related problems, we can use a GNN-based predictor to learn system performance information from architecture graphs.
Therefore, inspired by HGNAS~\cite{zhou2023hardware}, GCoDE integrates GNN predictors to assess the performance of GNN architectures on target device-edge co-inference systems.

\subsection{Architecture Graph Construction}

Unlike the performance awareness of GNNs on a single device, predictor learning faces more challenges in heterogeneous device-edge co-inference systems.
To achieve accurate awareness across various system configurations, GCoDE introduces an enhanced node feature generation method.
The architecture graph abstraction and the enhanced node feature generation are detailed as follows.

\textbf{Graph abstraction.}
For a GNN architecture sampled from the unified design space, GCoDE first abstracts it into a directed acyclic graph.
In this graph, nodes represent various operations, while edges indicate data flow between operations.
To further enhance the connectivity of the architecture graph, GCoDE introduces global nodes to connect all other nodes and adds self-connections.

\textbf{Enhanced node features generation.}
For GNN-based predictors, node features construction of the architecture graph commonly uses a one-hot encoding approach, as seen in HGNAS~\cite{zhou2023hardware}.
However, unlike the single-device focus of previous work, the device-edge hierarchies are generally heterogeneous.
This indicates that identical operation in the same architecture graph might show vastly different execution characteristics.
Therefore, the straightforward application of one-hot encoding strategies fails to effectively incorporate heterogeneous information, potentially resulting in poor performance of GNN-based predictors (See Sec.~\ref{sec:predictorResult}).
To better capture system heterogeneity and network conditions, GCoDE combines lookup table (LUT) and predictor methods to construct enhanced node features for the architecture graph.
Specifically, GCoDE maintains an LUT that records the performance of operations on different devices, with the overhead of constructing this lookup table being essentially negligible due to the limited number of valid operations.
Subsequently, GCoDE concatenates the one-hot encoding of each node's type with its corresponding performance from the LUT, enhancing the information provided by the initial node features.
As a result, the node features will comprise two components: operation type features and operation performance features.
To mitigate the impact of different operation magnitudes, performance values in the LUT are normalized using \textit{z-score normalization} before concatenation.
In this manner, the initial node features in the architecture graph incorporate vital information about heterogeneous performance and network conditions, which support the effective learning of the GNN predictor.

\subsection{Latency Prediction}

With the system configuration, architecture graph, enhanced node features, and maintained operation latency LUT, GCoDE develops a GNN-based predictor model to estimate the architecture's latency on the target device-edge system.
Since the architecture graph constructed by GCoDE contains sufficient system performance information, latency estimation can be considered a process of extracting information from the graph.
As such, GCoDE builds the latency predictor based on three GIN layers.
Additionally, each GIN layer uses a \textit{mean} aggregation operator to aggregate latency information across nodes.
Furthermore, \textit{Global Sum Pooling} is applied at the tail of the predictor to further capture latency information aggregated by the GIN layers.
The combination of the \textit{mean} aggregation operator and \textit{Global Sum Pooling} allows the predictor to efficiently extract latency information from the entire graph.
Consequently, our latency predictor achieves high prediction accuracy, ensuring that the explored architectures meet the strict latency requirements of edge application scenarios.
Additionally, since the architecture graph contains very few nodes, the predictor's runtime overhead is in the millisecond range, making it negligible.

In practice, incorrectly predicting high-latency architectures as low-latency can lead to substandard design.
To avoid such mispredictions, GCoDE employs a LUT estimation approach to correct the predictor outputs.
Specifically, using the latency LUT maintained by GCoDE, we estimate the architecture latency by simply accumulating all the operation latency in the corresponding architecture graph.
Although it may not account for all potential runtime overheads, the estimated latency provides a lower bound.
Thus, if the predicted result is lower than the estimated result, it indicates an incorrect prediction.
In such cases, GCoDE corrects the prediction by adopting the LUT estimation as the evaluation result.

\subsection{On-Device Energy Consumption Prediction}

In real-world edge applications, the energy consumption on devices during co-inference is a crucial performance metric.
Particularly for mobile devices, on-device energy consumption during inference determines the feasibility of the designed solution.
To achieve the energy awareness, most existing work relies on traditional estimation methods~\cite{odema2021lens}, the same ones applied in our previous conference publications.
Note that for device-edge co-inference, we focus primarily on the energy consumption of the device, as its energy is often limited.
In this estimation approach, the total energy consumption of the device for a single co-inference process can be calculate as: 
\begin{equation*}
    E_{total} = E_{idle} + E_{run} + E_{comm}\ ,
\end{equation*}
where $E_{comm}$ represents the communication energy consumption, computed using power models proposed in \cite{huang2012close}. 
$E_{idle}$ and $E_{run}$ are computational energy consumption in idle and operation executed states, respectively, calculated by multiplying associated power consumption with idle and execution time.

\begin{figure}[t]
    \centering
    \includegraphics[width = 0.9\linewidth]{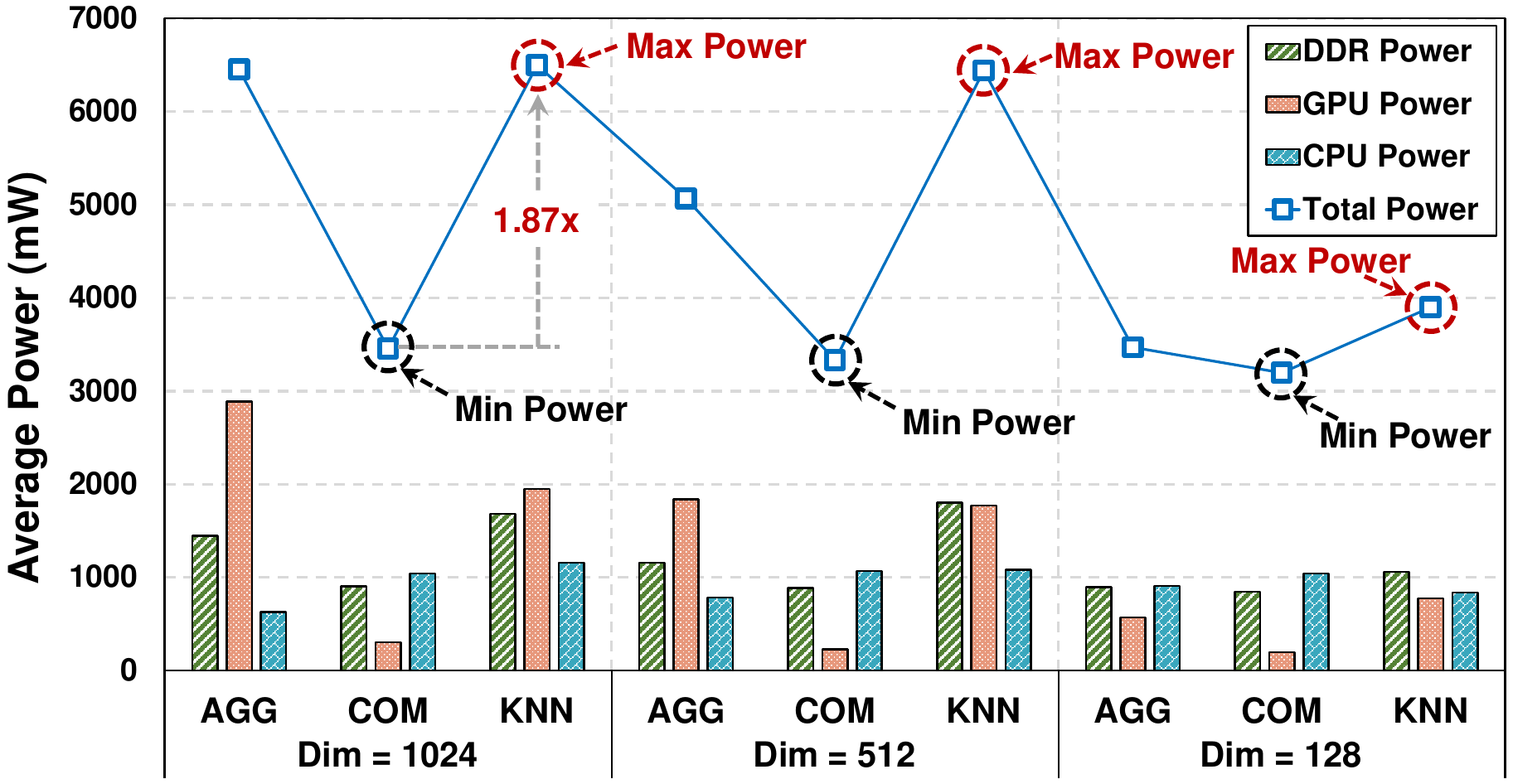}
    \caption{Execution power of GNN operations on Jetson TX2.}
    \label{fig:power}
    \vspace{-6pt}
\end{figure}

Those estimation methods assumes that the device's execution power remains constant throughout the co-inference process.
However, as demonstrated in Observation 2, the execution characteristics of GNNs differ across various operations.
This raises the question: Is it appropriate to use a fixed runtime power to estimate the energy consumption of various GNN architectures?
Fig.~\ref{fig:power} illustrates the profiling results of execution power for different GNN operations.
It is clear that there are significant differences in the average power during the execution of various operations.
For large feature dimensions, the \textit{KNN} operation requires $1.87\times$ more average power than the \textit{Combine} operation.
This is due to the significantly increased DDR and GPU load resulting from the intensive memory accesses required by \textit{KNN} operations.
Additionally, the difference in execution power consumption between operations becomes more pronounced as the feature size increases.
Moreover, the design solutions within the unified GNN design space may encompass a variety of different operation settings.
Thus, using a fixed runtime power multiplied by execution time to calculate the energy consumption for each sampled GNN architecture is not reasonable.
A fine-grained on-device energy awareness method is warranted to accurately consider the energy consumption differences among GNN operations.

To meet energy consumption requirements, GCoDE designs an on-device energy predictor to evaluate the energy usage of edge devices by candidate architectures during co-inference.
We randomly sampled $9000$ GNN architectures to gather on-device average power and latency data during co-inference, constructing the training dataset for the energy predictor.
For the measurement of the average execution power on Jetson TX2, we leveraged the sensing circuitry integrated on the device.
The energy consumption of each architecture is calculated by multiplying its latency by its average power.
Additionally, we construct an operation energy LUT to record the execution energy consumption for various GNN operations.
Following the construction of the training dataset, the energy predictor is trained for $500$ epochs, utilizing \textit{mean absolute percentage error} (MAPE) as the loss function.

\begin{figure}[t]
    \centering
    \includegraphics[width = 0.9\linewidth]{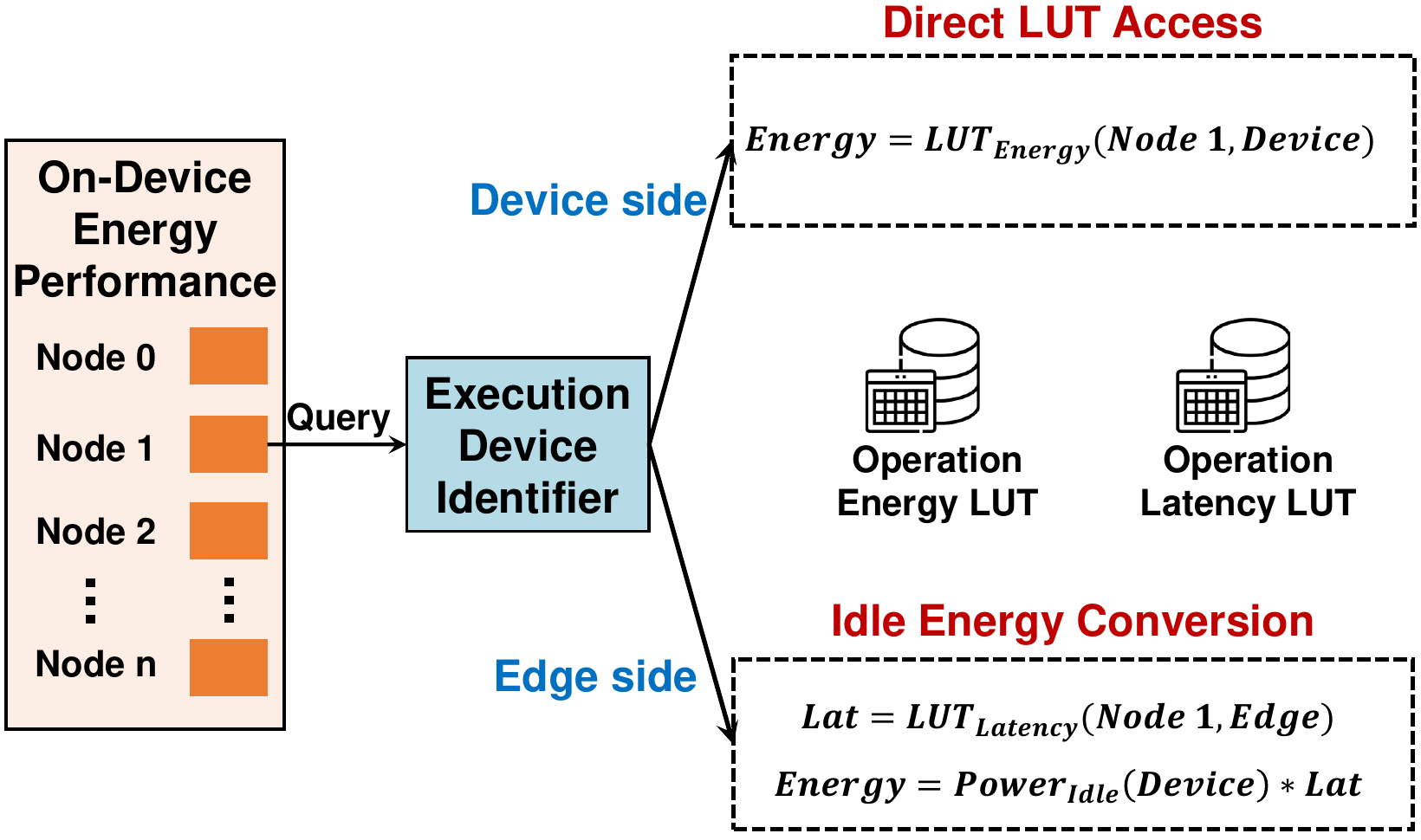}
    \caption{Specialized performance feature construction for each node in energy predictor.}
    \label{fig:energy_feature}
    \vspace{-6pt}
\end{figure}

As an extension of the system-aware approach, the model structure of the energy predictor is identical to that of the latency predictor.
However, unlike the latency predictor, the energy consumption predictor requires a specialized design for constructing enhanced node features.
This is because not all operations are executed on the device during co-inference.
Therefore, directly querying the LUT for performance features, as done in the latency predictor, is not applicable to the energy consumption predictor.
Additionally, simply summing the energy consumption of all operations fails to reflect the lower bound of on-device energy consumption.
To this end, we developed a specialized method for constructing performance features for the energy predictor, as illustrated in Fig.~\ref{fig:energy_feature}.
Specifically, when generating the performance feature part for each node, we first identified which side the node was executing on.
For operation nodes executed on the device side, features were obtained directly by querying the operation energy LUT.
While for the operation nodes executed on the edge side, we converted their energy consumption.
The energy consumption of these nodes is estimated by multiplying the device’s idle power consumption by the node’s execution time.
In this manner, the energy predictor effectively evaluates on-device energy consumption during GNN co-inference, enabling GCoDE to explore the most energy-efficient designs and meet energy consumption constraints.

\section{Device-Edge Deployment}\label{sec:deployment}

After identifying optimal GNN-Mapping pairings, GCoDE utilizes an integrated device-edge deployment approach to efficiently execute co-inference tasks on target systems.
The three key components are the GNN architecture zoo, the runtime dispatcher, and the pipelined co-inference engine.

\textbf{GNN architecture zoo.}
To address changing edge requirements, GCoDE maintains a collection of optimal architectures.
This is achieved by leveraging the integrated constraint-based random search strategy, enabling simultaneous exploration of multiple optimization objectives within a single search process.
Thus, we can identify the optimal architectures for varying requirements and system configurations in a single search process without extra overhead.
In this way, GCoDE can quickly adapt its deployment architecture to ensure efficient inference under varying system environments.

\begin{figure}[t]
    \centering
    \includegraphics[width = 0.8\linewidth]{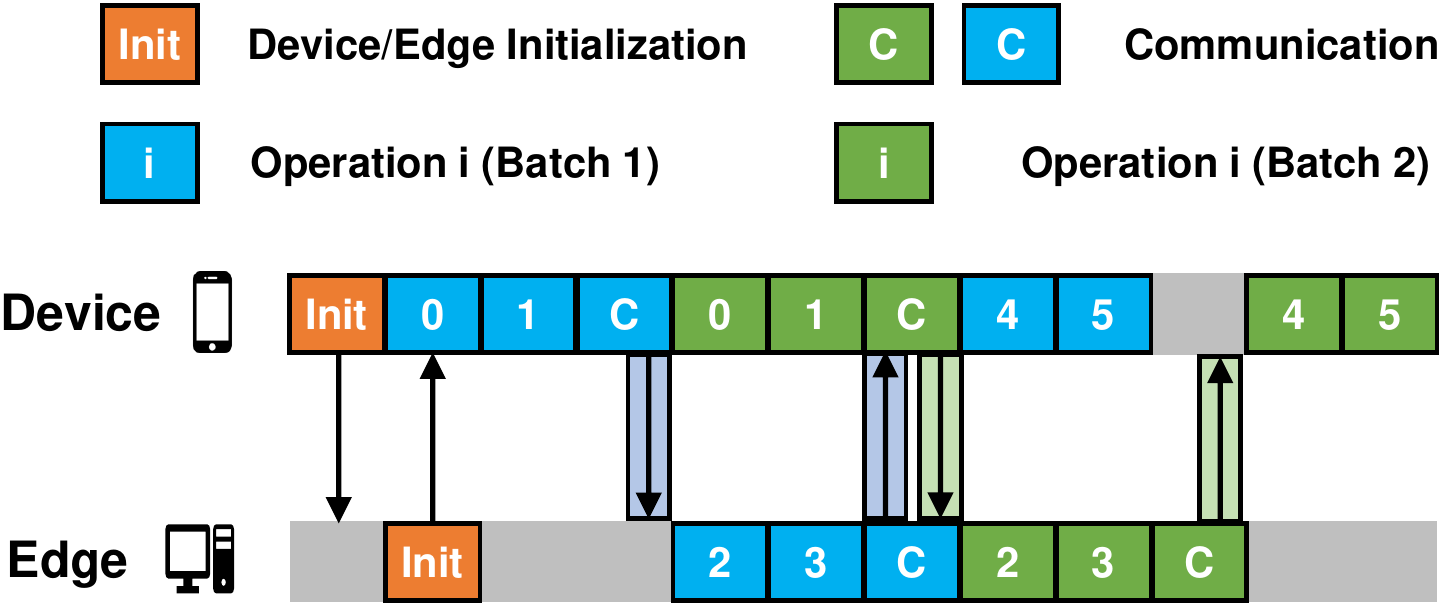}
    \caption{Device-edge co-inference pipeline of two batches, with numbers in cells indicating operation IDs.}
    \label{fig:pipeline}
    \vspace{-12pt}
\end{figure}

\textbf{Runtime dispatcher.}
Leveraging the maintained GNN architecture zoo, GCoDE allows for rapid switching of deployment solutions during runtime.
Specifically, GCoDE can choose the optimal GNN architecture for co-inference deployment according to user input or changes in the system environment.
To allow flexibility in switching deployment architectures, GCoDE does not pre-assign GNN architectures to the devices and the edge.
During each inference task execution, the edge device transmits the deployment architecture information to the edge server.
As a result, the deployment architecture can be adjusted dynamically in response to the demands of edge devices.

\textbf{Pipelined co-inference engine.}
There is a noticeable gap in operation-level fine-grained co-inference engines for GNNs, as most existing work is restricted to inter-layer collaboration.
For this purpose, GCoDE develops a GNN-specific pipelined co-inference engine leveraging Python Socket~\cite{socket}.
Fig.~\ref{fig:pipeline} shows the device-edge co-inference pipeline of two batches, effectively boosting system throughput.
Specifically, the device continues processing the next batch of data without idling while waiting for the edge to return intermediate results after executing the device-side operations of the deployed architecture.
When network conditions degrade, the overlapping computation time can adequately cover communication delays, enhancing overall efficiency.
Additionally, the co-inference engine utilizes multi-threading, with the sender and receiver operating independently, each maintaining its message queue.
To minimize communication overhead, GCoDE serializes and compresses messages during each communication process.
Moreover, GCoDE exhibits strong adaptability to heterogeneous environments owing to its modular communication design. In the prototype implementation, Python Socket is employed for rapid development and cross-platform validation, while the communication module can be readily replaced with alternative communication libraries to accommodate diverse deployment scenarios.

\section{Experiment}\label{sec:experiment}

\subsection{Experimental Settings}\label{sec:setting}

\begin{figure*}[t]
    \centering
    \includegraphics[width = 1\linewidth]{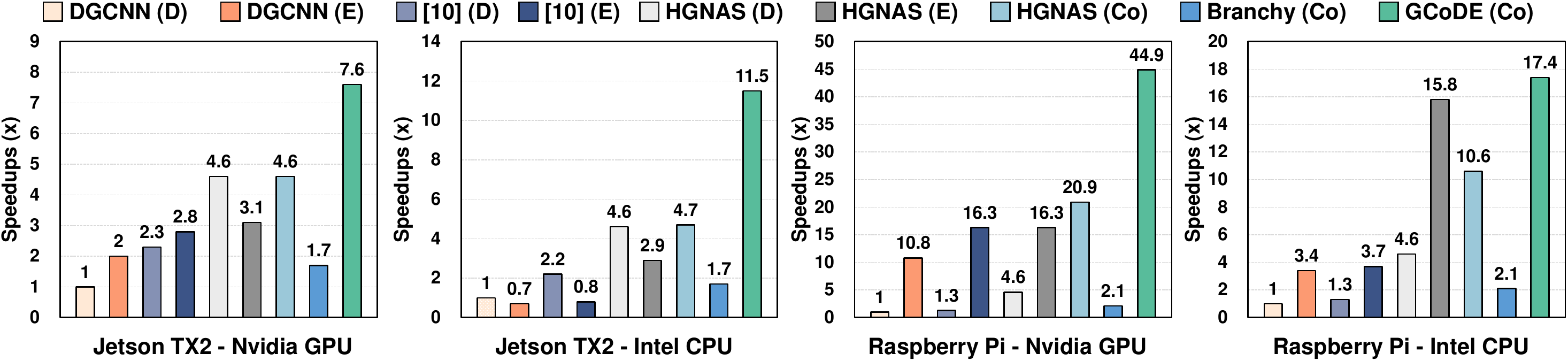}
    \caption{The inference speedups achieved by our GCoDE framework over eight competitors, using DGCNN as the baseline.}
    \label{fig:speedups}
    \vspace{-3pt}
\end{figure*}

\textbf{Datasets and competitors settings.}
Our evaluation considers \textbf{two different application datasets}: the point cloud processing dataset ModelNet40 \cite{wu20153d} and the text processing dataset MR \cite{ZhangYCWWW20}.
To evaluate GCoDE's performance in device-edge co-inference, we compare it against \textbf{five baselines}: (1) the manually designed DGCNN model~\cite{wang2019dynamic}, (2) a manually optimized GNN architecture derived from DGCNN~\cite{li2021towards}, (3) Branchy~\cite{shao2021branchy}, a GNN device-edge co-inference approach leveraging model splitting, (4) PAS~\cite{wei2023neural}, a GNN NAS framework focused on graph classification tasks, and (5) HGNAS~\cite{zhou2023hardware}, a hardware-aware GNN NAS framework for edge devices.
Additionally, to demonstrate that GCoDE's improvement in inference efficiency is not dependent on the computational power of the edge server, we established \textbf{three inference modes} for different competitors.
Specifically, Device-Only (D) denotes that the inference task is performed entirely on the edge devices, Edge-Only (E) denotes that the data is sent to edge servers for inference, and Co-Inference (Co) denotes that the GNN inference task is distributed and executed across the device-edge hierarchy.
Both GCoDE and Branchy, as co-inference methods, are executed in Co-Inference mode.
Furthermore, we partitioned hardware-efficient GNNs from existing NAS frameworks and evaluated their efficiency at optimal split points to highlight the limitations of separating architecture and mapping design.
For hyperparameter settings, we use the same setup as HGNAS for point cloud experiments and follow PAS for text processing.
For a fair comparison, we used the reported task accuracy in these papers and evaluated efficiency based on the PyTorch Geometric (PyG) framework \cite{Fey/Lenssen/2019} under the same experimental conditions.

\textbf{Device-edge system configurations.}
To compare the efficiency of GCoDE and competitors, we employ \textbf{four device-edge configurations}: Jetson TX2~\cite{tx2} and Raspberry Pi 4B~\cite{raspberry} as the device, and Nvidia 1060 GPU~\cite{1060} and Intel i7-7700 CPU~\cite{cpu} as the edge.
All devices are connected to a wireless router, with varying network conditions simulated by setting upload bandwidth limits ($\mathcal{S}_{L}$) to $10$ Mbps and $40$ Mbps.
Furthermore, to ensure a fair comparison, all competitors use the same system configuration and co-inference engine as GCoDE in their experiments.
Additionally, all implementations and tests are conducted based on the PyG framework, and the reported system performance results are from actual measurements on the target systems.
The on-device energy consumption of Jetson TX2 is measured using its integrated sensing circuits, while the Raspberry Pi estimates running energy consumption following \cite{odema2021lens} since it lacks integrated sensing circuits.

\textbf{Implementation settings for GCoDE.}
For the unified GNN design space, GCoDE built a 12-layer supernet and selected the functions from DGCNN as the initial function setting.
During the GNN-mapping co-exploration, the operation search runs for up to $1000$ iterations.
Given the search efficiency and the architecture after the operational search were adequate for real-time applications, the function scale-down tuning process was omitted.
For the latency and energy consumption predictors, we collected performance data from $9000$ randomly sampled GNN architectures within the unified GNN design space to construct the training dataset ($70\%/30\%$ for training/validation).
Additionally, \textit{mean absolute percentage error} (MAPE) is used as the loss function for training the predictor.

\begin{table*}[ht]
\centering
\caption{Performance comparison of GCoDE and existing approaches in different modes: Device-Only (D), Edge-Only (E), and Device-Edge Co-Inference (Co). OA and mAcc denote overall and balanced accuracy, respectively. Lat. and En. denote latency and on-device energy, respectively.}
\label{tab:mn_compare}
\renewcommand\arraystretch{1.4}
\resizebox{\linewidth}{!}{%
\begin{tabular}{|c|c|c|c|c|cccc|cccc|}
\hline
\multirow{3}{*}{Edge} &
  \multirow{3}{*}{Mode} &
  \multirow{3}{*}{Method} &
  \multirow{3}{*}{OA} &
  \multirow{3}{*}{mAcc} &
  \multicolumn{4}{c|}{Device: Jetson TX2} &
  \multicolumn{4}{c|}{Device: Raspberry Pi 4B} \\ \cline{6-13} 
 &
   &
   &
   &
   &
  \multicolumn{2}{c|}{$\mathcal{S}_{L}$$\le$ 40 Mbps} & 
  \multicolumn{2}{c|}{$\mathcal{S}_{L}$$\le$ 10 Mbps} &
  \multicolumn{2}{c|}{$\mathcal{S}_{L}$$\le$ 40 Mbps} &
  \multicolumn{2}{c|}{$\mathcal{S}_{L}$$\le$ 10 Mbps} \\ \cline{6-13} 
 &
   &
   &
   &
   &
  \multicolumn{1}{c|}{Lat. (ms)} &
  \multicolumn{1}{c|}{En. (J)} &
  \multicolumn{1}{c|}{Lat. (ms)} &
  En. (J) &
  \multicolumn{1}{c|}{Lat. (ms)} &
  \multicolumn{1}{c|}{En. (J)} &
  \multicolumn{1}{c|}{Lat. (ms)} &
  En. (J) \\ \hline
\multirow{3}{*}{-} &
  \multirow{3}{*}{D} &
  DGCNN~\cite{wang2019dynamic} &
  92.9 &
  88.9 &
  \multicolumn{1}{c|}{241.9} &
  \multicolumn{1}{c|}{1.0} &
  \multicolumn{1}{c|}{241.9} &
  1.0 &
  \multicolumn{1}{c|}{1121.8} &
  \multicolumn{1}{c|}{5.6} &
  \multicolumn{1}{c|}{1121.8} &
  5.6 \\
 &
   &
  \cite{li2021towards} &
  92.6 &
  90.6 &
  \multicolumn{1}{c|}{107.6} &
  \multicolumn{1}{c|}{0.4} &
  \multicolumn{1}{c|}{107.6} &
  0.4 &
  \multicolumn{1}{c|}{851.1} &
  \multicolumn{1}{c|}{4.3} &
  \multicolumn{1}{c|}{851.1} &
  4.3 \\ 
   &
    &
  HGNAS~\cite{zhou2023hardware} &
  92.1$\sim$92.2 &
  88.3$\sim$88.7 &
  \multicolumn{1}{c|}{52.1} &
  \multicolumn{1}{c|}{0.2} &
  \multicolumn{1}{c|}{52.1} &
  0.2 &
  \multicolumn{1}{c|}{241.5} &
  \multicolumn{1}{c|}{1.2} &
  \multicolumn{1}{c|}{241.5} &
  1.2 \\ \hline
\multirow{6}{*}{\parbox{0.8cm}{\centering Nvidia\\GPU}} &
  \multirow{3}{*}{E} &
  DGCNN~\cite{wang2019dynamic} &
  92.9 &
  88.9 &
  \multicolumn{1}{c|}{118.8} &
  \multicolumn{1}{c|}{0.2} &
  \multicolumn{1}{c|}{123.9} &
  0.3 &
  \multicolumn{1}{c|}{103.5} &
  \multicolumn{1}{c|}{0.4} &
  \multicolumn{1}{c|}{107.8} &
  0.4 \\ 
 &
   &
  \cite{li2021towards} &
  92.6 &
  90.6 &
  \multicolumn{1}{c|}{86.6} &
  \multicolumn{1}{c|}{0.2} &
  \multicolumn{1}{c|}{93.4} &
  0.2 &
  \multicolumn{1}{c|}{68.7} &
  \multicolumn{1}{c|}{0.2} &
  \multicolumn{1}{c|}{75.8} &
  0.3 \\ 
 &
   &
  HGNAS~\cite{zhou2023hardware} &
  92.1 &
  88.5 &
  \multicolumn{1}{c|}{79.2} &
  \multicolumn{1}{c|}{0.2} &
  \multicolumn{1}{c|}{87.8} &
  0.2 &
  \multicolumn{1}{c|}{69.0} &
  \multicolumn{1}{c|}{0.2} &
  \multicolumn{1}{c|}{70.3} &
  0.3 \\ \cline{2-13} 
 &
  \multirow{3}{*}{Co} &
  Branchy~\cite{shao2021branchy} &
  92 &
  - &
  \multicolumn{1}{c|}{141.2} &
  \multicolumn{1}{c|}{0.6} &
  \multicolumn{1}{c|}{141} &
  0.6 &
  \multicolumn{1}{c|}{541.8} &
  \multicolumn{1}{c|}{2.6} &
  \multicolumn{1}{c|}{531.8} &
  2.6 \\ 
 &
   &
  HGNAS~\cite{zhou2023hardware} &
  92.1$\sim$92.2 &
  88.3$\sim$88.7 &
  \multicolumn{1}{c|}{52.6} &
  \multicolumn{1}{c|}{0.2} &
  \multicolumn{1}{c|}{57.1} &
  0.2 &
  \multicolumn{1}{c|}{53.7} &
  \multicolumn{1}{c|}{1.9} &
  \multicolumn{1}{c|}{72.9} &
  0.9 \\ 
 &
   &
  \textbf{GCoDE} &
  \textbf{92.1$\sim$92.8} &
  \textbf{88.1$\sim$89.7} &
  \multicolumn{1}{c|}{\textbf{31.9}} &
  \multicolumn{1}{c|}{\textbf{0.1}} &
  \multicolumn{1}{c|}{\textbf{39}} &
  \textbf{0.1} &
  \multicolumn{1}{c|}{\textbf{25.0}} &
  \multicolumn{1}{c|}{\textbf{0.1}} &
  \multicolumn{1}{c|}{\textbf{35.6}} &
  \textbf{0.1} \\ \hline
\multirow{6}{*}{\parbox{0.8cm}{\centering Intel\\CPU}} &
  \multirow{3}{*}{E} &
  DGCNN~\cite{wang2019dynamic} &
  92.9 &
  88.9 &
  \multicolumn{1}{c|}{347.6} &
  \multicolumn{1}{c|}{0.8} &
  \multicolumn{1}{c|}{350.1} &
  0.8 &
  \multicolumn{1}{c|}{333.7} &
  \multicolumn{1}{c|}{1.0} &
  \multicolumn{1}{c|}{339.5} &
  1.1 \\ 
 &
   &
  \cite{li2021towards} &
  92.6 &
  90.6 &
  \multicolumn{1}{c|}{321.6} &
  \multicolumn{1}{c|}{0.7} &
  \multicolumn{1}{c|}{325.5} &
  0.8 &
  \multicolumn{1}{c|}{303.4} &
  \multicolumn{1}{c|}{1.0} &
  \multicolumn{1}{c|}{307.7} &
  1.0 \\ 
 &
   &
  HGNAS~\cite{zhou2023hardware} &
  92.1 &
  88.5 &
  \multicolumn{1}{c|}{83.7} &
  \multicolumn{1}{c|}{0.2} &
  \multicolumn{1}{c|}{88.3} &
  0.2 &
  \multicolumn{1}{c|}{71.0} &
  \multicolumn{1}{c|}{0.3} &
  \multicolumn{1}{c|}{74.0} &
  0.3 \\ \cline{2-13} 
 &
  \multirow{3}{*}{Co} &
  Branchy~\cite{shao2021branchy} &
  92 &
  - &
  \multicolumn{1}{c|}{140.2} &
  \multicolumn{1}{c|}{0.6} &
  \multicolumn{1}{c|}{140.8} &
  0.6 &
  \multicolumn{1}{c|}{528.1} &
  \multicolumn{1}{c|}{2.5} &
  \multicolumn{1}{c|}{544.0} &
  2.6 \\ 
 &
   &
  HGNAS~\cite{zhou2023hardware} &
  92.1$\sim$92.2 &
  88.3$\sim$88.7 &
  \multicolumn{1}{c|}{51.4} &
  \multicolumn{1}{c|}{0.2} &
  \multicolumn{1}{c|}{53.8} &
  0.2 &
  \multicolumn{1}{c|}{106.0} &
  \multicolumn{1}{c|}{2.1} &
  \multicolumn{1}{c|}{122.8} &
  1.0 \\ 
 &
   &
  \textbf{GCoDE} &
  \textbf{92.0$\sim$92.6} &
  \textbf{88.9$\sim$89.4} &
  \multicolumn{1}{c|}{\textbf{21}} &
  \multicolumn{1}{c|}{\textbf{0.1}} &
  \multicolumn{1}{c|}{\textbf{50.2}} &
  \textbf{0.2} &
  \multicolumn{1}{c|}{\textbf{64.4}} &
  \multicolumn{1}{c|}{\textbf{0.2}} &
  \multicolumn{1}{c|}{\textbf{49.3}} &
  \textbf{0.2} \\ \hline
\end{tabular}%
}
\end{table*}

\subsection{Evaluation on Point Cloud Processing}\label{sec:point_eval}

\textbf{Overall performance.}
Fig.~\ref{fig:speedups} illustrates the speedup comparison of GCoDE against all other SOTA baselines.
GCoDE achieves the most significant inference efficiency improvements across all system configurations, with speedups of $7.6\times$, $11.5\times$, $44.9\times$, and $17.4\times$, respectively.
The acceleration of GCoDE is even more significant with the Jetson TX2 - Intel CPU and Raspberry Pi - Nvidia GPU system configurations.
This phenomenon aligns with the hardware sensitivities noted in our earlier observation (Observation 2).
Given the differing execution bottlenecks of edge devices and edge servers in these two configurations, GCoDE effectively recognizes and utilizes this heterogeneity to achieve greater performance improvements.

\textbf{GCoDE vs. Existing approaches.}
Tab.~\ref{tab:mn_compare} provides detailed experimental results, including comparisons of accuracy, latency, and on-device energy consumption.
Compared to directly deploying manually designed DGCNN models on target edge devices, GCoDE achieves up to $44.9\times$ speedups and $98.2\%$ energy savings while maintaining similar accuracy.
Against hardware-efficient GNNs designed by HGNAS for edge devices, GCoDE consistently shows optimal performance, achieving up to $9.7\times$ speedup and $91.6\%$ energy savings on the lower-powered Raspberry Pi.
The much higher energy reduction ratio compared to latency reduction is due to the large performance gap between the Raspberry Pi and the edge server, which leads the optimizer to offload most operations to the server. As our measured energy consumption only accounts for on-device energy, this results in a relatively greater energy reduction than latency reduction.
Furthermore, compared to these efficient GNN architectures operating in Edge-Only mode, GCoDE still provides up to $5.7\times$ acceleration and over $50\%$ energy savings.
This demonstrates that GCoDE effectively leverages the potential of the device-edge co-inference paradigm to achieve optimal system performance.
Compared to the GNN device-edge co-inference approach Branchy, GCoDE achieves up to $21.9\times$ speedups and $96.1\%$ energy savings.
Additionally, GCoDE outperforms Branchy across all tests, even under varying network conditions and device-edge hierarchies.
This can be attributed to GCoDE's ability to jointly optimize the GNN architecture and operation mapping while leveraging system performance awareness to explore optimal designs.
Compared to HGNAS with its best partitioning point selection, GCoDE still achieves up to $2.5\times$ inference speedups.
This highlights that the separation of architecture design and operation mapping leads to sub-optimal performance, while GCoDE’s co-optimization approach maximizes system performance.

\begin{figure}[t]
    \centering
    \includegraphics[width = 0.85\linewidth]{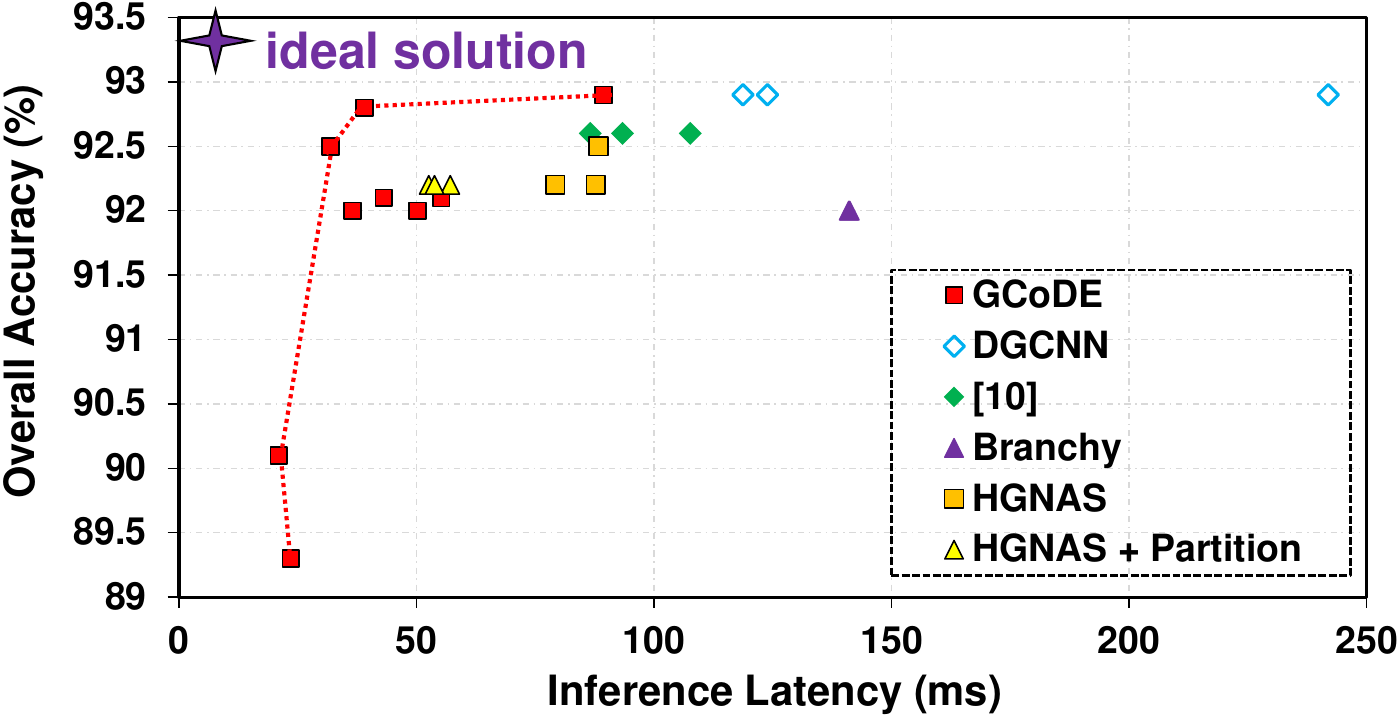}
    \caption{Comparison between the existing approaches and GCoDE in terms of accuracy and latency.}
    \label{fig:pareto}
    \vspace{-9pt}
\end{figure}

\textbf{Accuracy vs. efficiency.}
Fig.~\ref{fig:pareto} illustrates the GNN design space exploration results using Jetson TX2 as the device.
GCoDE advances the Pareto frontier in GNN inference performance beyond all baselines, achieving both higher accuracy and lower latency.
This advancement is due to our system performance predictor, which identifies efficient GNN architectures, bringing GCoDE closer to the ideal solution. Furthermore, the selection of scaling factor $\lambda$ in the search process allows users to tailor the GNN co-inference architecture for either higher accuracy (smaller $\lambda$) or lower latency (larger $\lambda$), based on their needs.

\subsection{Evaluation on Text Processing}

\begin{table}[t]
\centering
\caption{Comparison of existing methods and GCoDE on MR dataset.}
\renewcommand\arraystretch{1.5}
\resizebox{1.0\linewidth}{!}{%
\begin{tabular}{|ll|c|c|cc|c|}
\hline
\multicolumn{2}{|l|}{}                                           & \textbf{GCoDE} & Branchy~\cite{shao2021branchy} & \multicolumn{2}{c|}{PAS~\cite{wei2023neural}} & \multicolumn{1}{l|}{PAS~\cite{wei2023neural}} \\ \hline
\multicolumn{2}{|c|}{Accuracy (\%)}                             & \textbf{76.1$\sim$77.0} & 75.5 & \multicolumn{2}{c|}{76.7} & 76.7  \\ \hline
\multicolumn{2}{|c|}{Mode}                                      & \textbf{Co}          & Co   & D           & E           & Co    \\ \hline
\multicolumn{1}{|l|}{\multirow{2}{*}{TX2 $\rightleftharpoons$ GPU}} & Lat. (ms) & \textbf{8.7}   & 26.4  & 29.1        & 30.7        & 16.2                               \\
\multicolumn{1}{|l|}{}                           & En. (mJ)   & \textbf{25.3}        & 78.3  & 94.3         & 71.4         & 49.3   \\ \hline
\multicolumn{1}{|l|}{\multirow{2}{*}{TX2 $\rightleftharpoons$ CPU}}  & Lat. (ms) & \textbf{8.5}         & 29.0   & 29.1        & 18.6     & 15.5 \\
\multicolumn{1}{|l|}{}                           & En. (mJ)   & \textbf{24.5}        & 87.9  & 94.3         & 45.4         & 47.2   \\ \hline
\multicolumn{1}{|l|}{\multirow{2}{*}{Pi $\rightleftharpoons$ GPU}} & Lat. (ms) & \textbf{4.8}         & 32.3 & 13.6        & 32.0          & 8.0  \\
\multicolumn{1}{|l|}{}                           & En. (mJ)   & \textbf{30.0}        & 150.0  & 70.0         & 140.0         & 41.0  \\ \hline
\multicolumn{1}{|l|}{\multirow{2}{*}{Pi $\rightleftharpoons$ CPU}}   & Lat. (ms) & \textbf{2.00}           & 28.7 & 13.6        & 28.7        & 6.9  \\
\multicolumn{1}{|l|}{}                           & En. (mJ)   & \textbf{10.0}        & 140.0  & 70.0         & 110.0         & 37.0   \\ \hline
\end{tabular}%
}
\label{tab:mr}
\end{table}

To evaluate the performance of GCoDE across different applications, we conducted experiments on the MR text analysis dataset under a $40$ Mbps network condition.
The experimental results are illustrated in Tab.~\ref{tab:mr}, which demonstrate that GCoDE consistently maintains superior accuracy and system efficiency compared to all competitors.
Specifically, GCoDE outperforms PAS, which is deployed in Device-Only mode, achieving speedups of $3.3\times$, $3.4\times$, $2.8\times$, and $6.8\times$ across four heterogeneous system configurations.
Compared to the architecture designed by PAS and deployed in Edge-Only mode, GCoDE achieves $3.5\times$, $2.2\times$, $6.7\times$, and $14.4\times$ improvements in inference efficiency across four system configurations, respectively.
Against co-inference methods like Branchy and PAS, which utilize optimal partitioning points, GCoDE achieves up to $14.3\times$ speedup.
Furthermore, GCoDE is the most energy-efficient approach among all baselines, requiring only $10$ mJ on the Raspberry Pi for a single inference. 

\begin{figure}[t]
    \centering
    \includegraphics[width = 1\linewidth]{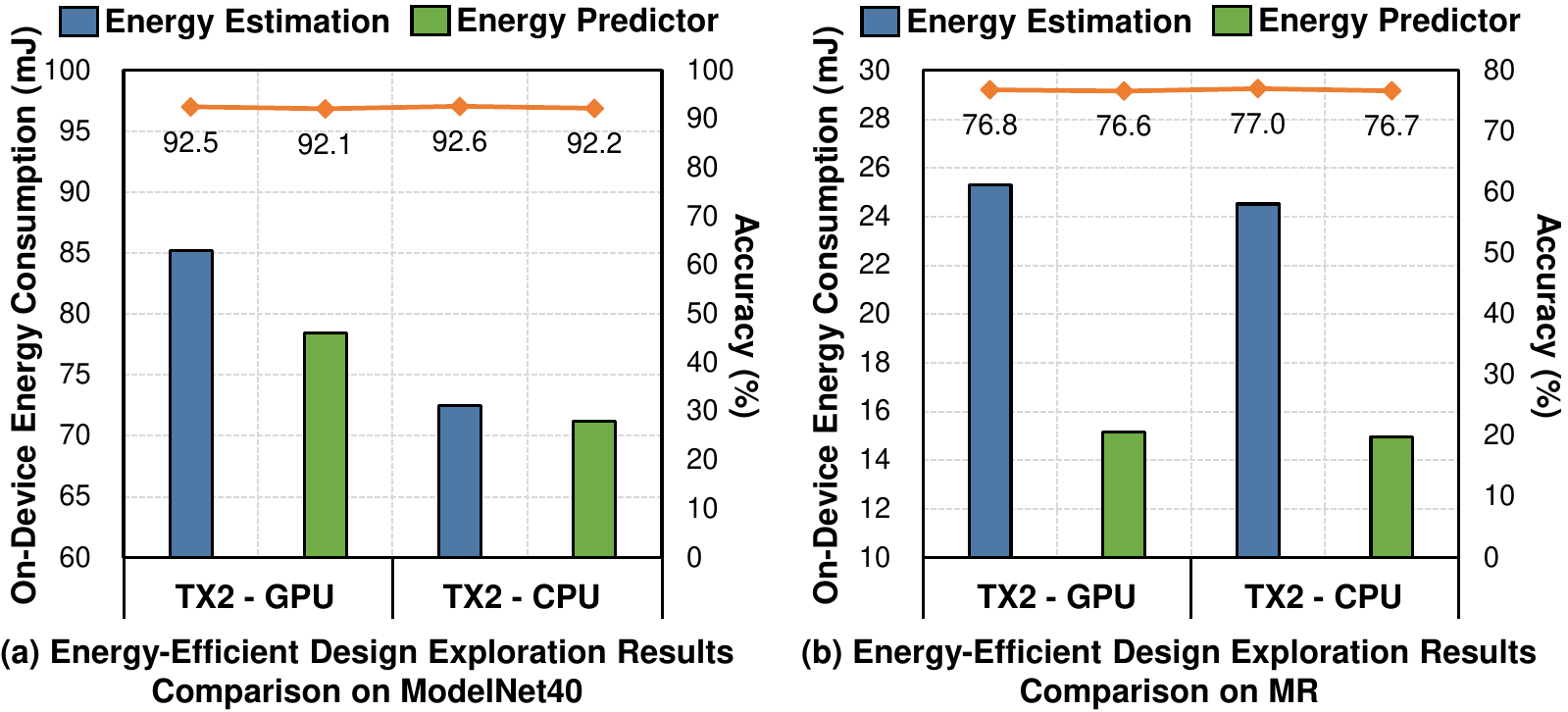}
    \caption{Performance comparison of exploration based on energy predictor or energy estimation.}
    \label{fig:energy_search}
    \vspace{-6pt}
\end{figure}

\subsection{Energy-Efficient Design Exploration Results}

Fig.~\ref{fig:energy_search}~(a) shows the results of energy-efficient design exploration on the ModelNet40 dataset.
During the exploration, the on-device energy consumption of the candidate architectures is evaluated using the traditional estimation method and our predictor, respectively.
With a more accurate energy predictor, GCoDE effectively explores GNN architectures that consume less on-device energy during co-inference.
Specifically, the prediction method achieves up to $8\%$ reduction in energy consumption on point cloud processing.
Furthermore, Fig.~\ref{fig:energy_search}~(b) illustrates that with energy prediction methods, GCoDE achieves a further reduction in on-device energy consumption by $40\%$ and $39\%$ for text processing tasks, respectively.
In practice, GCoDE-designed architectures aimed at faster inference often show lower on-device energy consumption, aligning with the findings of \cite{luo2022surgenas}.
Thus, even though the energy estimation method is not accurate, architectures with relatively low energy consumption can still be identified by considering the impact of latency on the objective function.
Nonetheless, the effectiveness of architectures based on energy estimation methods remains uncertain, making them unsuitable for scenarios with strict energy constraints.
Conversely, the energy predictor provides outputs closer to the measured values, ensuring that the designed solution adheres to strict energy constraints.

\begin{figure}[t]
    \centering
    \includegraphics[width = 0.8\linewidth]{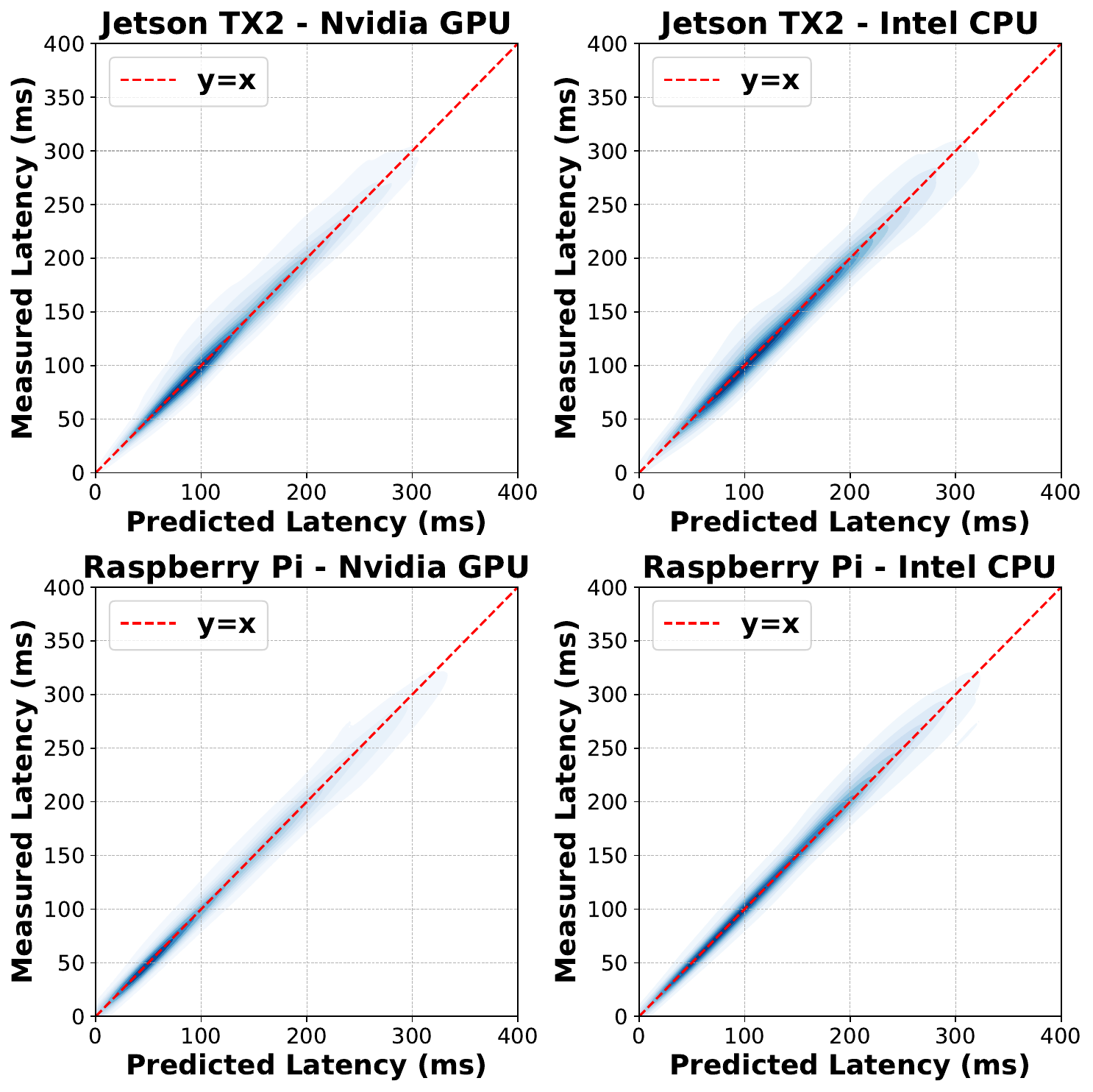}
    \caption{Illustration of the relationships between predicted and measured latency across four device-edge systems.}
    \label{fig:scatter}
\end{figure}

\begin{figure}[t]
    \centering
    \includegraphics[width = 0.9\linewidth]{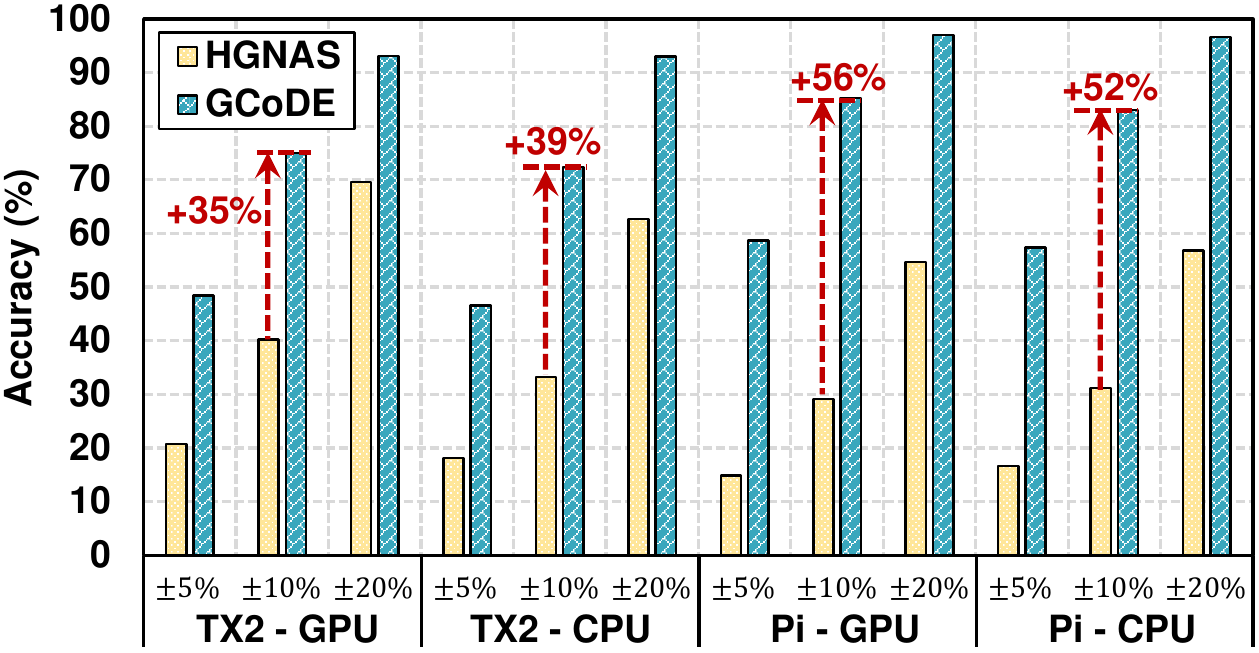}
    \caption{Latency prediction accuracy within 20\% error bound on four device-edge systems.}
    \label{fig:predictor_result_lat}
    \vspace{-9pt}
\end{figure}

\begin{figure}[t]
    \centering
    \includegraphics[width = 0.8\linewidth]{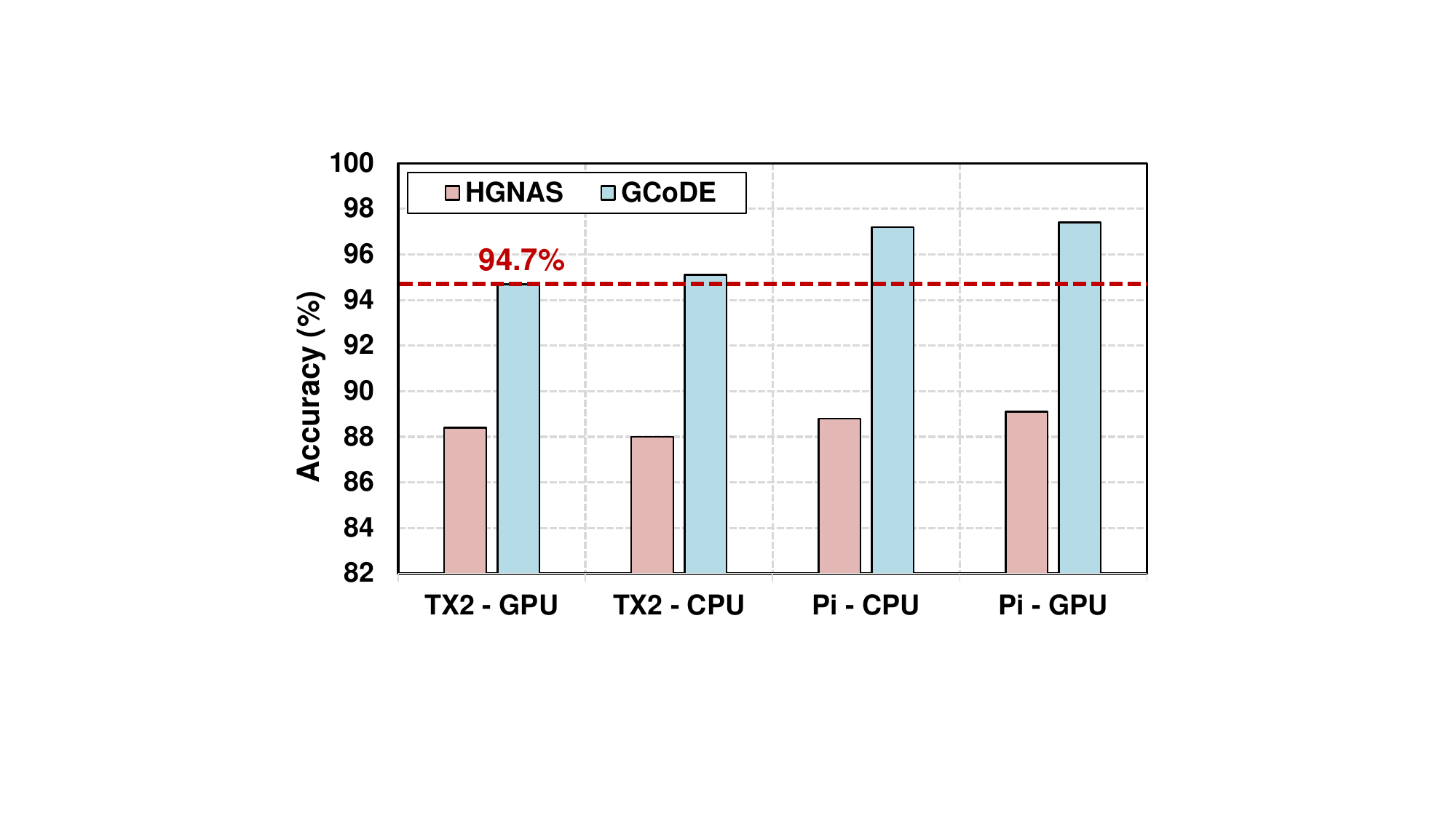}
    \caption{Prediction accuracy of relative latency ranking among candidate architectures.}
    \label{fig:rlpa}
\end{figure}

\subsection{System Performance Awareness Results}\label{sec:predictorResult}

In this section, we evaluate the performance of GCoDE in predicting latency and on-device energy consumption across four different device-edge system configurations.
Meanwhile, we extend the GNN hardware performance predictor proposed by HGNAS to enable awareness at the device-edge hierarchy, allowing for a fair comparison.

\textbf{Latency prediction.}
Fig.~\ref{fig:scatter} presents an intuitive illustration of the performance of our proposed latency predictor across different system configurations.
It is clear that our prediction results are very close to the measured co-inference latency on the target device-edge systems.
The average MAPE of latency predictions across the four device-edge hierarchies is approximately $6.9$.
Additionally, GCoDE demonstrates a notable improvement in accuracy for device-edge co-inference latency prediction over the HGNAS predictor, as illustrated in Fig.~\ref{fig:predictor_result_lat}.
Specifically, GCoDE achieves latency prediction accuracies of $75\%$, $72\%$, $85\%$, and $83\%$ across various system configurations within a $10\%$ error bound.
Compared to HGNAS, GCoDE achieves a prediction accuracy improvement of up to $52\%$ in GNN device-edge co-inference latency.
This is because HGNAS relies on a simple one-hot encoding strategy, and the node features in the architecture graph lack essential heterogeneous performance information.
In contrast, the enhanced node feature construction method introduced by GCoDE effectively incorporates system performance and heterogeneity information into the architecture graph, significantly facilitating the learning of the GNN predictor.
Besides, Fig.~\ref{fig:rlpa} further shows that GCoDE accurately evaluates the relative latency among candidate architectures, surpassing $94.7\%$ accuracy.
Moreover, to further assess the robustness of the proposed performance predictor, we evaluated it under diverse dynamic conditions, including varying network bandwidths (1, 20, 40, and 100 Mbps) and different edge workloads.
Using approximately 600 samples collected across these scenarios, the predictor achieved a relative latency ranking accuracy of $87.11\%$.
This high latency-aware performance is due to GCoDE's elegant abstraction of co-inference, which transforms complex system awareness problems into graph learning problems, allowing GNN predictors to solve them.
Furthermore, the latency prediction overhead is measured in milliseconds and is negligible.
By leveraging the latency predictor, GCoDE can efficiently identify the optimal GNN architecture with a mapping scheme for the target system.

\begin{figure}[t]
    \centering
    \includegraphics[width = 0.9\linewidth]{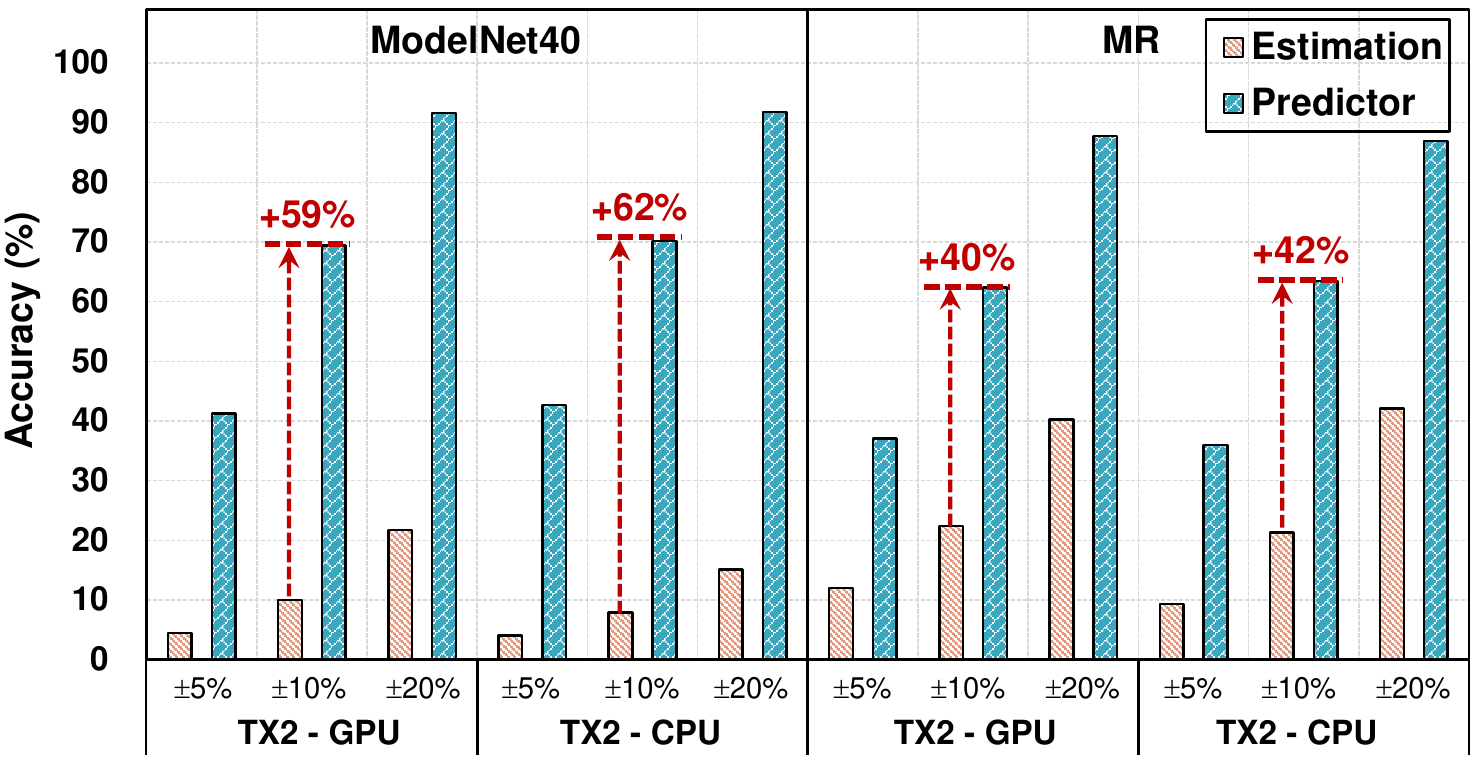}
    \caption{On-device energy consumption prediction accuracy within 20\% error bound on four device-edge systems.}
    \label{fig:energy_predict}
    \vspace{-9pt}
\end{figure}

\textbf{Energy prediction.}
Fig.~\ref{fig:energy_predict} illustrates the accuracy of the energy predictor in GCoDE.
Traditional estimation methods do not accurately reflect actual on-device energy consumption during co-inference, making it hard to meet energy constraints in practice.
By extending the predictor's awareness dimension, GCoDE achieves high accuracy in energy prediction through a specialized method for constructing the performance parts of node features.
Specifically, the energy predictor achieves prediction accuracies of $69\%$, $70\%$, $62\%$, and $63\%$ within a $10\%$ error bound across various system configurations and applications, respectively.
Compared to estimation methods, GCoDE achieves up to $62\%$ and $42\%$ improvements in energy consumption prediction accuracy for different applications, respectively.
Moreover, with a $20\%$ error bound, GCoDE can achieve over $87\%$ energy prediction accuracy.
By leveraging this accurate energy predictor, GCoDE can adhere to on-device energy consumption constraints of edge applications and identify more energy-efficient GNN solutions.

\begin{figure}[t]
    \centering
    \includegraphics[width = 1\linewidth]{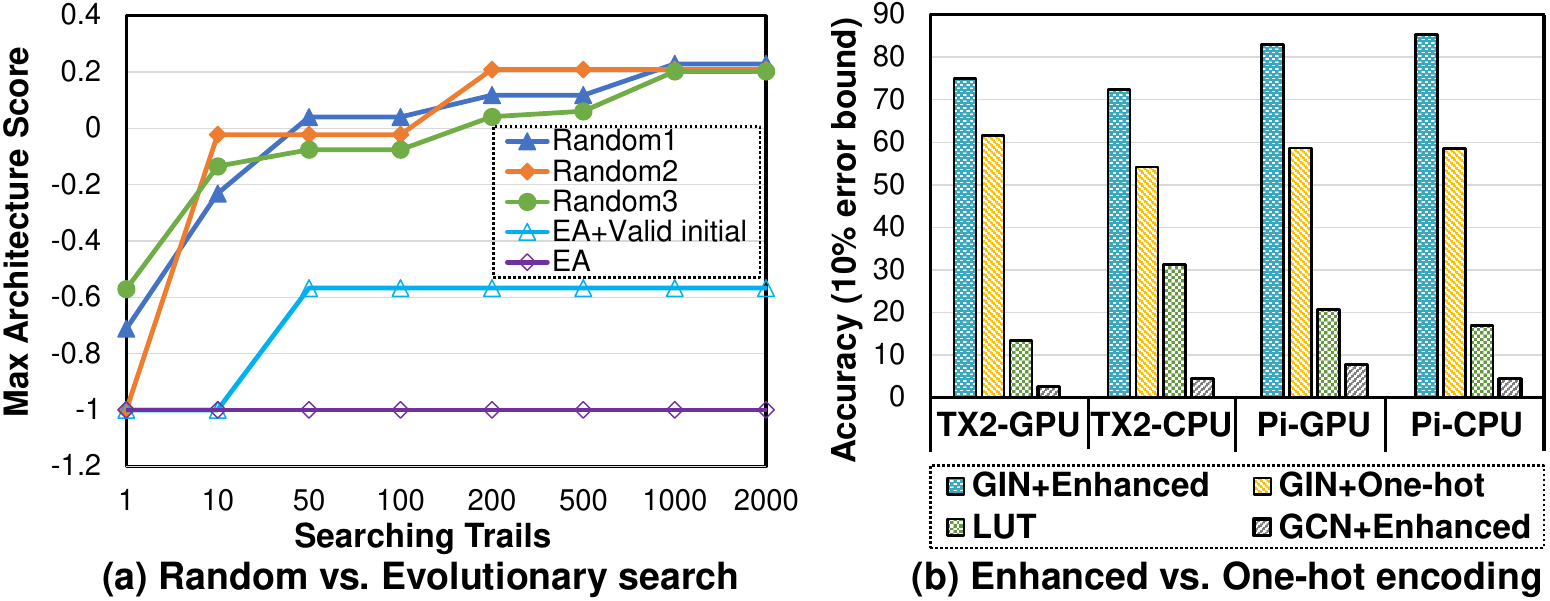}
    \caption{(a) Efficiency comparison between random search and evolutionary search. (b) Prediction accuracy improvement with GCoDE method (GIN + Enhanced feature).}
    \vspace{-9pt}
    \label{fig:ablation}
\end{figure}

\subsection{Ablation Studies}\label{sec:ablation}

In this section, we evaluate the effectiveness of the proposed constraint-based random search strategy in enhancing search efficiency, as well as the performance of various predictor construction methods.

\textbf{Random search vs. Evolutionary search.}
As shown in Fig.~\ref{fig:ablation}~(a), due to the presence of many invalid architectures in the unified GNN design space, the evolutionary search method gets stuck in a loop of identifying valid architectures, thereby losing the ability to search for higher-performing ones.
In contrast, the proposed constraint-based random search strategy performs remarkably well and can identify the optimal designs in fewer trials.
Specifically, GCoDE requires only 1.5 GPU hours to perform the search on the ModelNet40 dataset, doubling the exploration efficiency compared to the 3 GPU hours required by HGNAS.
For MR datasets with smaller sizes, GCoDE further reduces the search time to $0.2$ GPU hours, compared to PAS, which requires $0.27$ GPU hours.

\textbf{Enhanced feature construction vs. One-hot encoding strategy.}
Fig.~\ref{fig:ablation}~(b) shows a performance comparison of various prediction methods for device-edge co-inference.
Although the LUT method can evaluate performance lower bounds for GNN architectures, it struggles to achieve high prediction accuracy across different system configurations.
The reason is that it neglects essential runtime overhead.
Furthermore, GIN demonstrates a clear advantage over GCN in learning architecture graphs, owing to its superior graph information learning capability.
Moreover, while the powerful graph information extraction capability of GIN leads to an improvement in the accuracy of the one-hot encoding strategy, it remains ineffective.
In contrast, the combination of our enhanced node feature construction method and GIN's powerful learning capability leads to a significant improvement in prediction accuracy across various heterogeneous co-inference systems.

\section{Conclusions}\label{sec:conclusions}
In this paper, we propose GCoDE, the first system-aware automated framework for designing and deploying GNNs on device-edge co-inference systems.
Given user requirements, GCoDE can automatically design the optimal GNN architecture with an embedded operation mapping scheme, while providing efficient co-inference engine support.
GCoDE constructs a unified architecture-mapping co-design space, employing constraint-based search strategies and accurate system performance awareness approaches to identify optimal solutions.
Extensive experiments demonstrate that GCoDE achieves superior accuracy, inference speed, and energy efficiency across diverse applications and systems, surpassing baselines with up to $44.9\times$ acceleration and $98.2\%$ energy reduction.
We believe that GCoDE brings significant heuristic advances in deploying efficient GNNs for large-scale wireless network edge applications.

\bibliographystyle{IEEEtran}
\footnotesize
\begingroup
\bibliography{ref.bib}
\endgroup


\vspace{-15mm}

\begin{IEEEbiography}[{\includegraphics[width=1in,height=1.25in,clip,keepaspectratio]{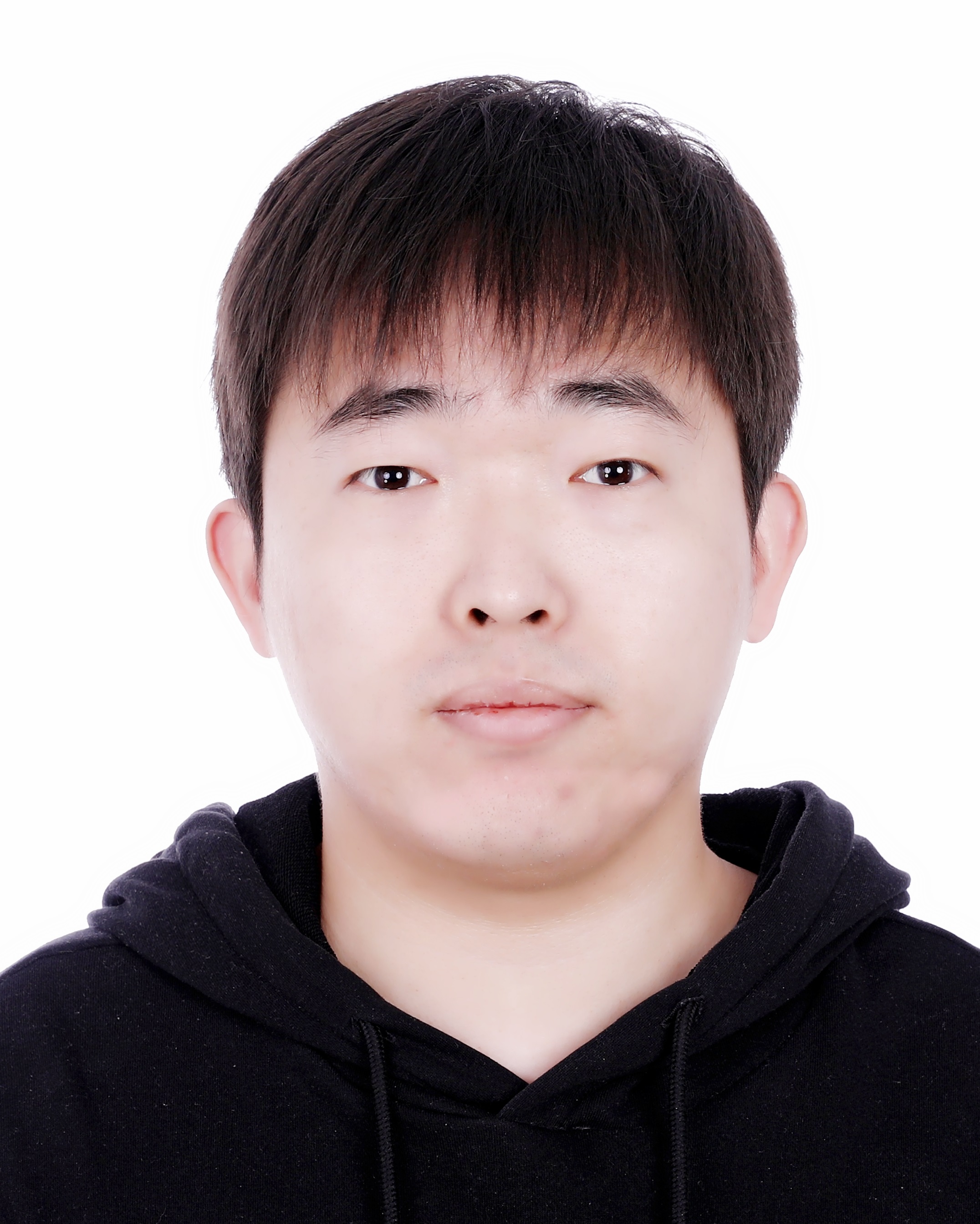}}]{Ao Zhou}

received the B.S. and M.S. degree in software engineering from Beijing University of Technology, Beijing, China, in 2018 and 2021, respectively, and the Ph.D. degree in software engineering from Beihang University, Beijing, China, in 2025. He is currently a Postdoctoral Researcher at the School of Software, Beihang University. His research interests include GNN acceleration, computer architecture, FPGA accelerator, and heterogeneous computing. He is one of the contributors to the popular GNN computation framework PyG.

\end{IEEEbiography}

\vspace{-20mm}

\begin{IEEEbiography}[{\includegraphics[width=1in,height=1.25in,clip,keepaspectratio]{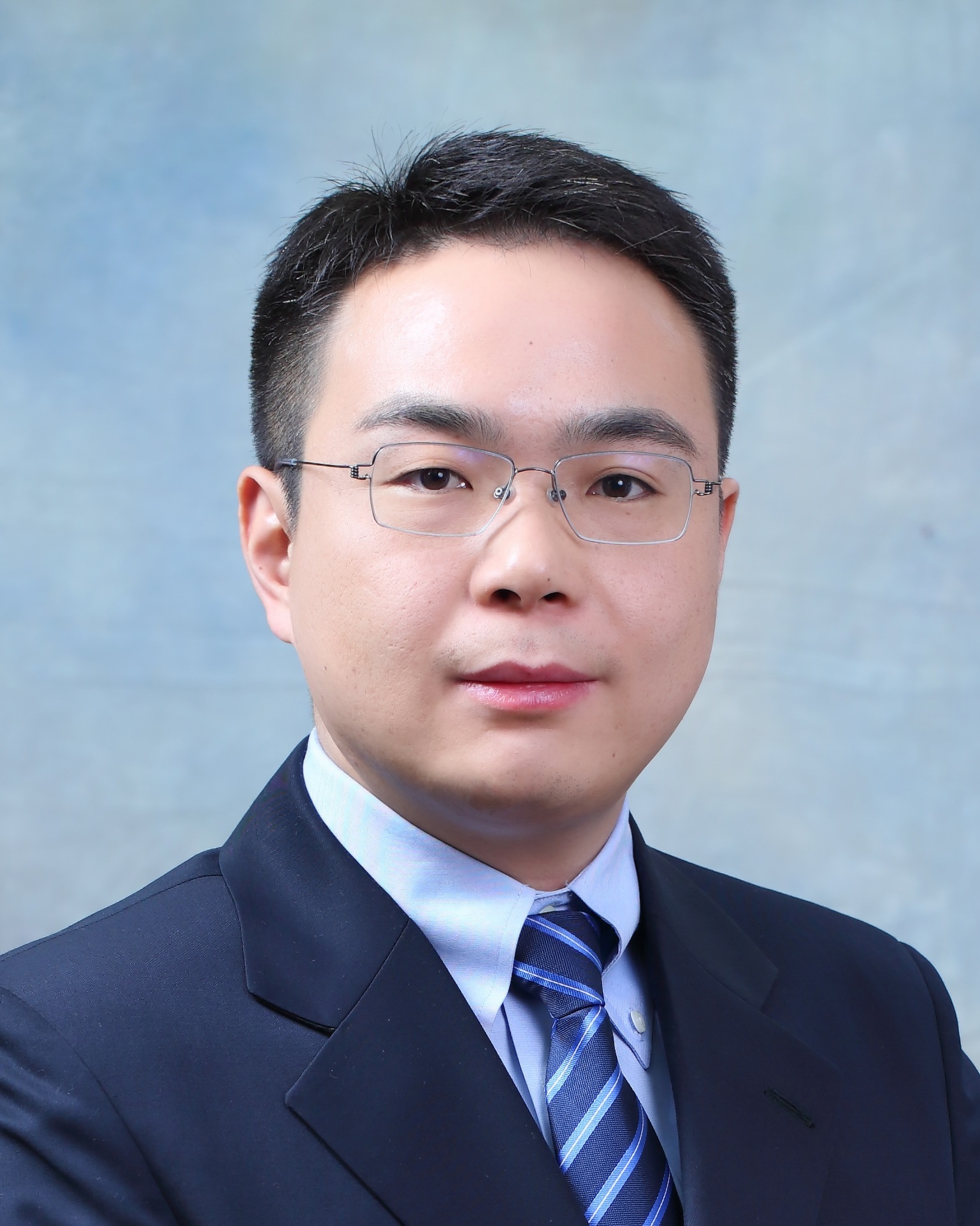}}]{Jianlei Yang}

(S'11-M'14-SM'20) received the B.S. degree in microelectronics from Xidian University, Xi'an, China, in 2009, and the Ph.D. degree in computer science and technology from Tsinghua University, Beijing, China, in 2014.

He is currently a Professor in Beihang University, Beijing, China, with the School of Computer Science and Engineering. From 2014 to 2016, he was a post-doctoral researcher with the Department of ECE, University of Pittsburgh, Pennsylvania, USA.
His current research interests include emerging computer architectures, hardware-software co-design and machine learning systems.

Dr. Yang was the recipient of the First/Second place on ACM TAU Power Grid Simulation Contest in 2011 and 2012. He was a recipient of IEEE ICCD Best Paper Award in 2013, ACM GLSVLSI Best Paper Nomination in 2015, IEEE ICESS Best Paper Award in 2017, ACM SIGKDD Best Student Paper Award in 2020.

\end{IEEEbiography}

\vspace{-16mm}

\begin{IEEEbiography}[{\includegraphics[width=1in,height=1.25in,clip,keepaspectratio]{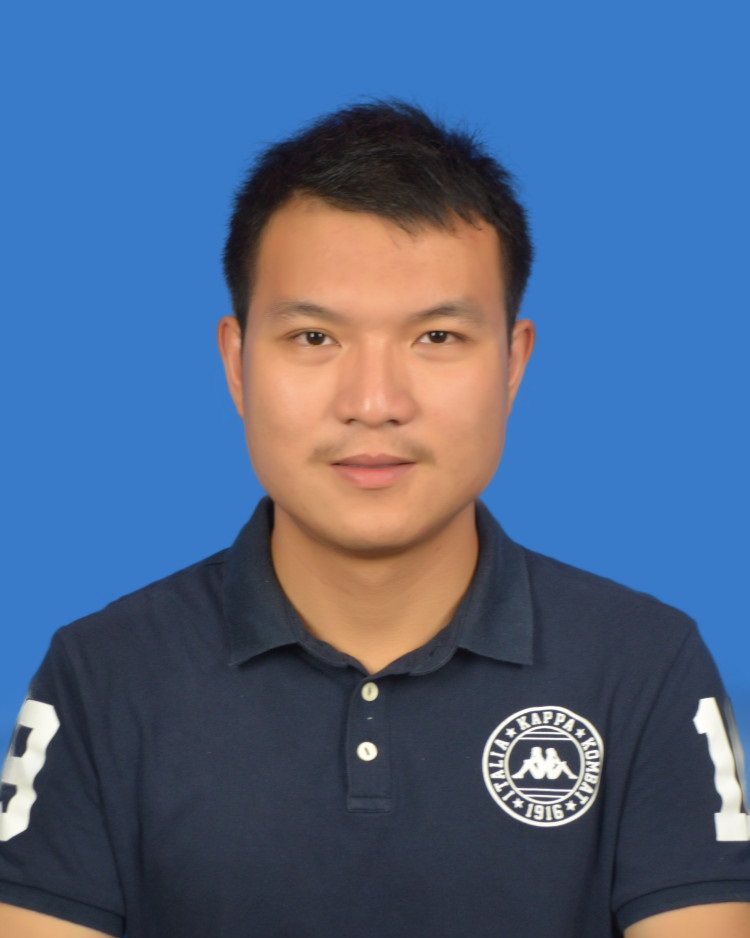}}]{Tong Qiao}

received the B.S. degree in computer science and technology from Beihang University, Beijing, China, in 2020. He is currently pursuing the Ph.D. degree at the School of Computer Science and Engineering, Beihang University, China. His research interests include graph neural networks acceleration, and system for machine learning.

\end{IEEEbiography}

\vspace{-12mm}

\begin{IEEEbiography}[{\includegraphics[width=1in,height=1.25in,clip,keepaspectratio]{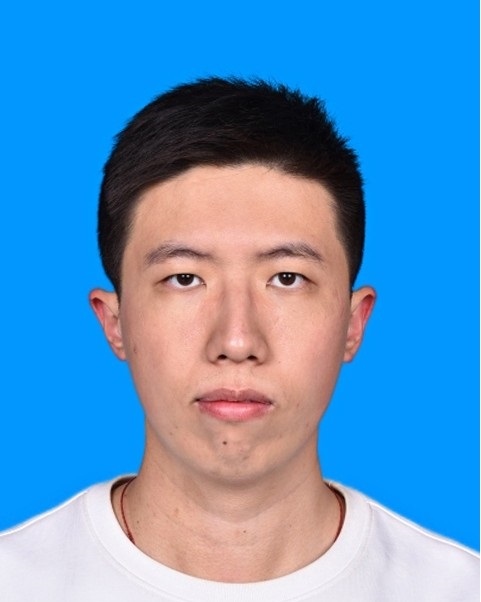}}]{Yingjie Qi}

received the B.S. degree in computer science and technology from Beihang University, Beijing, China, in 2020. He is currently pursuing the Ph.D. degree at the School of Computer Science and Engineering, Beihang University, China. His research interests include compute-in-memory architectures, deep learning compilers, and graph neural networks acceleration.

\end{IEEEbiography}

\vspace{-12mm}

\begin{IEEEbiography}[{\includegraphics[width=1in,height=1.25in,clip,keepaspectratio]{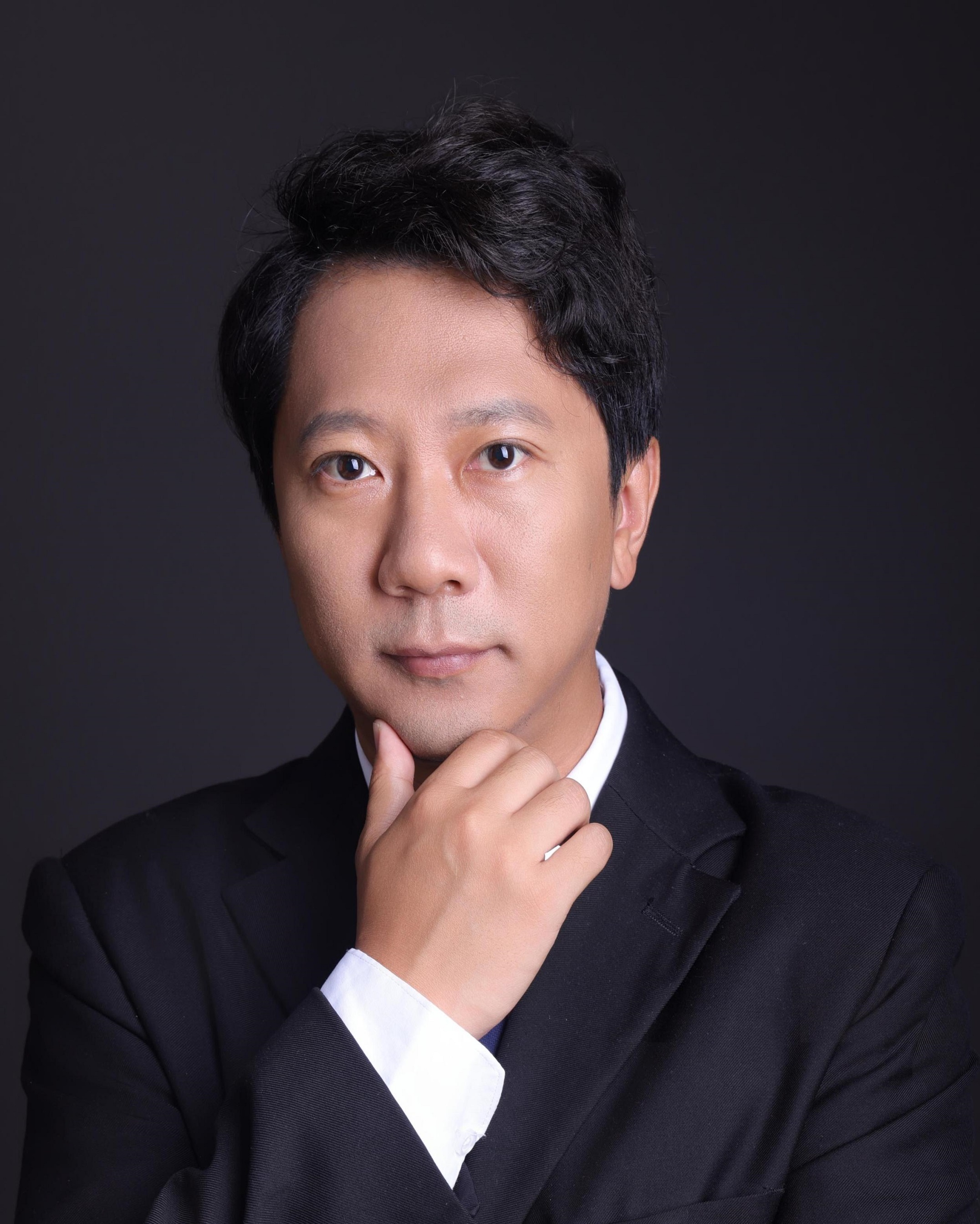}}]{Zhi Yang}

is currently an associate researcher in the School of Computer Science in Peking University. He obtained his Ph.D. from Peking University in 2010. His major research interests include AI computing systems and distributed systems. He was a recipient of VLDB Best Scalable Data Science Paper in 2022, WWW Best Student Paper Award in 2022.

\end{IEEEbiography}

\vspace{-16mm}
\begin{IEEEbiography}[{\includegraphics[width=1in,height=1.25in,clip,keepaspectratio]{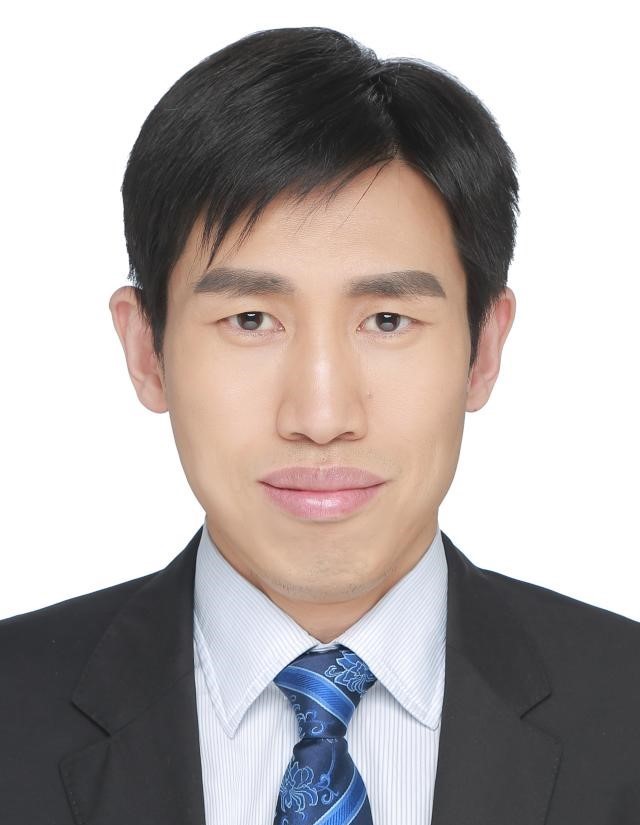}}]{Weisheng Zhao}

(Fellow, IEEE) received the Ph.D. degree in physics from the University of Paris Sud, Paris, France, in 2007.

He is currently a Professor with the School of Integrated Circuit Science and Engineering, Beihang University, Beijing, China. In 2009, he joined the French National Research Center, Paris, as a Tenured Research Scientist. Since 2014, he has been a Distinguished Professor with Beihang University. He has published more than 300 scientific articles in leading journals and conferences, such as \textit{Nature
Electronics}, \textit{Nature Communications}, \textit{Advanced Materials}, IEEE Transactions, ISCA, and DAC. His current research interests include the hybrid integration of nanodevices with CMOS circuit and new nonvolatile memory (40-nm technology node and below) like MRAM circuit and architecture design.

Prof. Zhao was the Editor-in-Chief for the {\sc{IEEE Transactions on Circuits and System I: Regular Paper}} from 2020 to 2023.

\end{IEEEbiography}

\vspace{-20mm}
\begin{IEEEbiography}[{\includegraphics[width=1in,height=1.25in,clip,keepaspectratio]{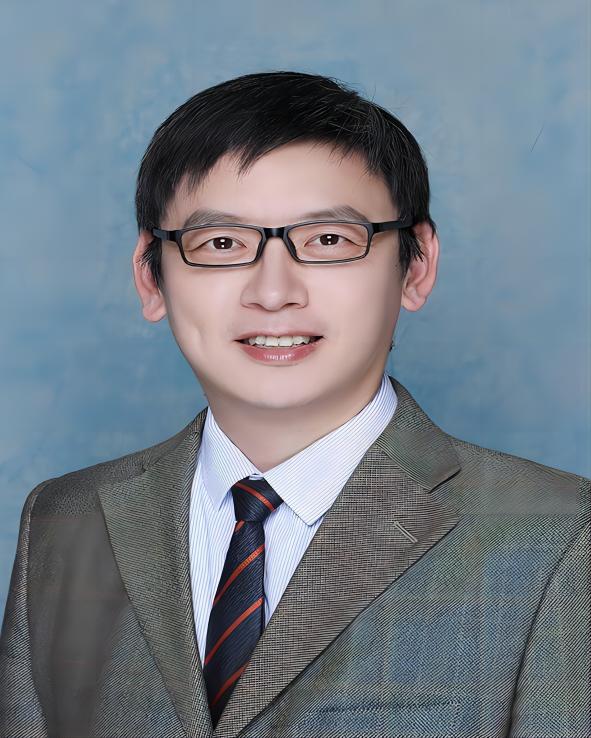}}]{Chunming Hu}

received the PhD degree in computer science and technology from Beihang University, Beijing, China, in 2006.

He is currently a professor and dean of the School of Software, Beihang University, Beijing, China. His current research interests include distributed systems, system virtualization, mobile computing and cloud computing.

Prof. Hu is currently one of the W3C Board of Directors, and serving as the Deputy Director of W3C China.

\end{IEEEbiography}

\end{document}